%% file: paper.tex
\documentclass[10pt,conference]{IEEEtran}
\IEEEoverridecommandlockouts

\include{macro}

\def\BibTeX{{\rm B\kern-.05em{\sc i\kern-.025em b}\kern-.08em
    T\kern-.1667em\lower.7ex\hbox{E}\kern-.125emX}}

\begin{document}   
 
\title{Fairify: Fairness Verification of Neural Networks}

%%%% Author information 
% \author{\IEEEauthorblockN{Anonymous Author(s)}} 

\author{\IEEEauthorblockN{Sumon Biswas}
\IEEEauthorblockA{\textit{School of Computer Science} \\
\textit{Carnegie Mellon University}\\
Pittsburgh, PA, USA \\
sumonb@cs.cmu.edu}
\and
\IEEEauthorblockN{Hridesh Rajan}
\IEEEauthorblockA{\textit{Dept. of Computer Science} \\
\textit{Iowa State University}\\
Ames, IA, USA \\
hridesh@iastate.edu}
}

\maketitle
% For page numbers
\thispagestyle{plain}
\pagestyle{plain}

\input{abstract}


% \begin{CCSXML}
% 	<ccs2012>
% 	<concept>
% 	<concept_id>10011007.10011074</concept_id>
% 	<concept_desc>Software and its engineering~Software creation and management</concept_desc>
% 	<concept_significance>500</concept_significance>
% 	</concept>
% 	<concept>
% 	<concept_id>10010147.10010257</concept_id>
% 	<concept_desc>Computing methodologies~Machine learning</concept_desc>
% 	<concept_significance>500</concept_significance>
% 	</concept>
% 	</ccs2012>
% \end{CCSXML}

% \ccsdesc[500]{Software and its engineering~Software creation and management}
% \ccsdesc[500]{Computing methodologies~Machine learning}

\input{introduction}
\input{preliminaries}
\input{approach}
\input{evaluation}
\input{threat}
\input{related}
\input{conclusion}

\renewcommand{\bibfont}{\footnotesize}
\bibliographystyle{IEEEtranN}
\bibliography{refs} 

% \appendix
% \input{appendix}

\end{document}

%% file: macro.tex
\usepackage{xcolor} % \usepackage[table]{xcolor}
\usepackage{multirow}
\usepackage{booktabs} % For formal tables
\usepackage{graphicx}
\usepackage[belowskip=-12pt]{caption}
\usepackage{rotating}
\usepackage{tabularx}
\usepackage{subfig}
\usepackage{wrapfig}
\usepackage{listings}
\usepackage{color, colortbl}
\usepackage{balance} % to balance the bibliography
\usepackage{lscape}
\usepackage[hidelinks]{hyperref}
\usepackage{xspace}
\usepackage{framed}
\usepackage{syntax}
\usepackage{soul}
\usepackage{flushend}
\usepackage{mathtools}% Loads amsmath
\usepackage{amsmath,amssymb,amsfonts}
\newtheorem{definition}{Definition}

\usepackage{algorithm}
\usepackage{algpseudocode} % noend
\usepackage{pifont}
\usepackage[clock]{ifsym}

\usepackage[tikz]{bclogo}
\usepackage[numbers,sort&compress]{natbib} %% for \citeauthor{}

\usepackage{hhline}
\usepackage{diagbox}
\usepackage{array, makecell}
\usepackage{microtype}

\usepackage{enumitem}
\definecolor{cellgreen}{RGB}{202,236,193}
\definecolor{cellblue}{RGB}{204, 229, 255}
\definecolor{cellpurple}{RGB}{216,191,216}
\definecolor{cellred}{RGB}{255, 204, 204}
\definecolor{cellorange}{RGB}{255, 229, 204}
\definecolor{cellgold}{RGB}{240,230,140}
\definecolor{cellsteel}{RGB}{230,230,250}

\definecolor{codegreen}{rgb}{0,0.6,0}
\definecolor{codegray}{rgb}{0.5,0.5,0.5}
\definecolor{codepurple}{rgb}{0.58,0,0.82}
\definecolor{codeblue}{rgb}{0, 0, 0.8}
\definecolor{backcolour}{rgb}{0.97,0.97,0.97}
\definecolor{Gray}{gray}{0.4}

\lstdefinestyle{mystyle}{
	backgroundcolor=\color{backcolour},   
	commentstyle=\color{Gray},
	keywordstyle=\color{codepurple},
	numberstyle=\tiny\color{codegray},
	stringstyle=\color{magenta},
	basicstyle=\tiny,
	breakatwhitespace=false,         
	breaklines=true,                 
	captionpos=b,                    
	keepspaces=true,                 
	numbers=left,                    
	numbersep=5pt,                  
	showspaces=false,                
	showstringspaces=false,
	showtabs=false,                  
	tabsize=2,
	xleftmargin=.015\textwidth, 
	otherkeywords={self, numpy.maximum, math.exp}
}

\lstdefinelanguage{Pythonna}{%
	language     = python,
	morekeywords = {to_categorical, flow_from_directory, pad_sequences, load_image, np.maximum}
}

\lstdefinestyle{customc}{
	belowcaptionskip=1\baselineskip,
	breaklines=false,
	frame= single,
	breaklines = true,
	xleftmargin=\parindent,
	language= Pythonna,
	showstringspaces=false,
	basicstyle=\footnotesize\ttfamily,
	keywordstyle=\bfseries\color{green!40!black},
	commentstyle=\itshape\color{purple!40!black},
	identifierstyle=\color{blue},
	stringstyle=\color{codegreen},
	backgroundcolor=\color{gray!4}
}

\graphicspath{{./figures/}}

\newcommand{\tabref}[1]{Table~\ref{#1}}
\newcommand{\fignref}[1]{Figure~\ref{#1}}
\newcommand{\algoref}[1]{Algorithm~\ref{#1}}

\newcommand{\secref}[1]{\S\ref{#1}}

\newcommand{\defref}[1]{Definition~\ref{#1}}

\newcounter{rqs}
\stepcounter{rqs}

\newcounter{NumObservations}
\stepcounter{NumObservations}
\definecolor{shadecolor}{rgb}{.9,.9,.9}

\newcommand*\circled[1]{\tikz[baseline=(char.base)]{
		\node[shape=circle,fill,inner sep=.6pt] (char) {\textcolor{white}{#1}};}}

%% file: abstract.tex
\begin{abstract}
% Background
% Gap
% Work

Fairness of machine learning (ML) software has become a major concern in the recent past. 
Although recent research on testing and improving fairness have demonstrated impact on real-world software, providing fairness guarantee in practice is still lacking.
Certification of ML models is challenging because of the complex decision-making process of the models.
In this paper, we proposed \textit{Fairify}, an SMT-based approach to verify individual fairness property in neural network (NN) models.
Individual fairness ensures that any two similar individuals get similar treatment irrespective of their protected attributes e.g., race, sex, age.
Verifying this fairness property is hard because of the global checking and non-linear computation nodes in NN.
We proposed sound approach to make individual fairness verification tractable for the developers. 
The key idea is that many neurons in the NN always remain inactive when a smaller part of the input domain is considered. So, Fairify leverages white-box access to the models in production and then apply formal analysis based pruning. Our approach adopts input partitioning and then prunes the NN for each partition to provide fairness certification or counterexample. 
We leveraged interval arithmetic and activation heuristic of the neurons to perform the pruning as necessary. We evaluated Fairify on 25 real-world neural networks collected from four different sources, and demonstrated the effectiveness, scalability and performance over baseline and closely related work. 
Fairify is also configurable based on the domain and size of the NN.
Our novel formulation of the problem can answer targeted verification queries with relaxations and counterexamples, which have practical implications. 
\end{abstract}

%% file: introduction.tex
\section{Introduction}
\label{sec:introduction}
Artificial intelligence (AI) based software are increasingly being used in critical decision making such as criminal sentencing, hiring employees, approving loans, etc. Algorithmic fairness of these software raised significant concern in the recent past \cite{aggarwal2019black,biswas20machine,biswas21fair,chakraborty2020fairway,hort2021fairea,chakraborty2021bias,zhang2021ignorance,zhang2020white}. Several studies have been conducted to measure and mitigate algorithmic fairness in software \cite{calders2010three,chouldechova2017fair,feldman2015certifying,hardt2016equality,dwork2012fairness,speicher2018unified,zafar2015fairness,zhang2018mitigating,pleiss2017fairness,zemel2013learning}. However, providing formal guarantee of fairness properties in practice is still lacking. 
Fairness verification in ML models is difficult given the complex decision making process of the algorithms and the specification of the fairness properties \cite{brun2018software,albarghouthi2017fairsquare,john2020verifying, bastani2019probabilistic}. Our goal in this paper is to enable the verification in real-world development and guarantee fairness in critical domains.

\citeauthor{albarghouthi2017fairsquare} and \citeauthor{bastani2019probabilistic} proposed probabilistic techniques to verify \textit{group fairness} \cite{albarghouthi2017fairsquare, bastani2019probabilistic}. 
% \citeauthor{bastani2019probabilistic} also proposed probabilistic verification technique for group fairness properties \cite{bastani2019probabilistic}. 
Group fairness property ensures that the protected groups (e.g., male-vs-female, young-vs-old, etc.) get similar treatment in the prediction.
On the other hand, \textit{individual fairness} states that any two similar individuals who differ only in their protected attribute get similar treatment \cite{galhotra2017fairness,john2020verifying}. 
\citeauthor{galhotra2017fairness} argued that group fairness property might not detect bias in scenarios when same amount of discrimination is made for any two groups \cite{galhotra2017fairness},
which led to the usage of individual fairness property in many recent works \cite{galhotra2017fairness,zhang2020white,aggarwal2019black,udeshi2018automated, zheng2021neuronfair}.
\citeauthor{john2020verifying} proposed individual fairness verification for two ML classifiers, i.e., linear classifier and 2) kernelized classifier e.g., support vector machine \cite{john2020verifying}, which is not applicable to neural networks. 

We propose \textit{Fairify}, the technique to verify individual fairness of NN models in production.
% \citeauthor{john2020verifying} proposed techniques to verify individual fairness for two classes of ML models: 1) linear classifier e.g., logistic regression, and 2) kernelized classifier e.g., support vector machine \cite{john2020verifying}. \citeauthor{zhang2020white} proposed individual fairness testing approach for NN \cite{zhang2020white}. \citeauthor{urban2020perfectly} proposed NN certification approach for \textit{dependency fairness} \cite{urban2020perfectly}.
% The authors leveraged abstract interpretation to compute input region that is certified as fair or unfair. The precision and scalability of the approach depend on the chosen abstract domains. 
Both abstract interpretation \cite{pulina2010abstraction,urban2020perfectly,mazzucato2021reduced, gehr2018ai2, singh2019abstract} and satisfiability modulo theories (SMT) based techniques \cite{huang2017safety,katz2017reluplex,katz2019marabou} have shown success in verifying different properties of NN such as robustness. However, fairness verification of NN on real-world models has received little attention.
We adopted SMT based verification since it enables practical benefits, e.g., solving arbitrary verification query and providing counterexamples.
\citeauthor{urban2020perfectly} proposed Libra, an abstract interpretation based \textit{dependency fairness} certification for NN \cite{urban2020perfectly}. The approach can not check relaxation of fairness queries and does not 
provide counterexamples in case of a violation.
Fairify on the other hand provides configurable options to the developers enabling fairness verification of NN in practice.

Verifying a property in NN is challenging mainly because of the presence of non-linear computation nodes i.e., activation functions \cite{katz2017reluplex,urban2020perfectly}. With the size of the NN, the verification task becomes harder and often untractable \cite{sun2018concolic, wang2018efficient}. 
Many studies have been conducted to verify NN for different local robustness properties \cite{gehr2018ai2,huang2017safety,ehlers2017formal}.
%such as robustness, collision avoidance, adversarial security, etc.
However, the individual fairness property requires global checking which makes the verification task even harder and existing local property verifiers can not be used \cite{gopinath2018deepsafe, katz2017towards}.
%Many of such techniques use satisfiability modulo theory (SMT) solvers to verify such properties under certain assumptions.  
To that end, we propose a novel technique that can verify fairness, i.e., provide satisfiability (SAT) with counterexample, or show UNSAT in a tractable time and available computational resource.
% is depicted in \fignref{fig:problem}, where 
Fairify takes a trained NN and the verification query as input. If the verifier can verify within the given timeout period, the output should be SAT (violation) or UNSAT (certification). When the verifier is unable to show any proof within the timeout, the result is UNK (unknown).
Our evaluation shows that using state-of-the-art SMT solver, we cannot verify the fairness property of the NN models in days; however, Fairify can verify most of the verification queries within an hour.
Fairify makes the verification task tractable and scalable by combining input partitioning and pruning approach.
%We further show that the approach is scalable to different relaxations of the fairness properties.

% \begin{figure}[!t]
% 	\centering
% 	\includegraphics[width=1\columnwidth]{problem.pdf}
% 	\caption{Fairness verification problem in NN}
% 	\label{fig:problem}
% \end{figure}

% \textbf{Input partitioning.} 
Our key insight is that the activation patterns of the NN used in practice are sparse when we consider only a smaller part of the input region.
%However, an arbitrary partitioning of input might not preserve the given fairness verification query. 
Therefore, we perform \textbf{input partitioning} to a certain point and divide the problem into multiple sub-problems which result into split-queries. To show the problem as UNSAT, all the split-queries have to be UNSAT. On the other hand, if any of the split-queries gives SAT, the whole problem becomes SAT.
Because of this construction of the problem, Fairify is able to provide certification or counterexample for each partition which has further value in fairness defect localization.

% \textbf{Sound pruning.} 
The input partitioning allowed us to perform static analysis on the network for the given partition and identify inactive neurons. We applied interval arithmetic to compute bounds of the neurons. In addition, we run individual verification query on each neuron to identify the ones that are always inactive. Since the inactive neurons do not impact the decision, removing those neurons gives a pruned version of the network that can be verified for the given partition. This \textbf{sound pruning} is lightweight, sound, and achieves high pruning ratio, which makes the verification task tractable.

% \textbf{Heuristic-based pruning.}
We further improve the efficiency of the approach by analyzing activation heuristics. We conduct lightweight simulation to profile the network and find candidate neurons that remain \textit{almost} always inactive but could not be removed through sound pruning. Thereby, we propose layer-wise \textbf{heuristics} that suggest inactive nodes; if necessary given the time budget. Although this pruning is based on heuristics, our evaluation shows that a conservative approach provides much improvement in the verification with negligible loss of accuracy.
Another novelty in this idea is that
% even if the heuristic based pruning might not result into a NN that always produces the same result as the original NN, 
the developer can choose to deploy the pruned (and verified) version of the NN. Thus, the pruned NN would provide sound fairness guarantee with little loss of accuracy.

After partitioning and reducing the complexity of the problem, we leverage a constraint solver to verify the split-queries on the pruned networks. Then we accumulate the results for each partition to provide verification for the original query.
We evaluated Fairify on 25 different NN models collected from four different sources. We collected appropriate real-world NNs from Kaggle \cite{kaggle} which are built for three popular fairness-critical tasks and took the NNs used in three prior works in the area \cite{zhang2020white,udeshi2018automated,urban2020perfectly}. Our results show significant improvement over the baseline with respect to utility, scalability, and performance. The main contributions of our work are as follows.

\begin{enumerate}[leftmargin=*]
    \item Fairify is the first to solve the individual fairness verification problem for already trained NN using SMT based technique.
	% We leveraged the white-box access to the NN in deployment and conducted lightweight analysis to prune it statically, which enabled the tractability of verification.
	
	\item Our formulation of the problem enables verification of different relaxation of the fairness property. Then we proposed two novel NN pruning methods designed to solve those queries effectively.
	% \hridesh{sound?}
	
	\item The approach can be integrated in the development pipeline and provides practical benefits for the developers i.e., certification or counterexample for each input partition and targeted fairness certification.
	
	\item We implemented Fairify using Python and openly available constraint solver Z3. We also created a benchmark of NN models for fairness verification. The code, models, benchmark datasets, and results are available in the self-contained GitHub repository 
	\cite{replication}
	\footnote{https://github.com/sumonbis/Farify} 
	that can be leveraged by future research.
\end{enumerate}

The rest of the paper is organized as follows: 
\secref{sec:background} describes the background and \secref{sec:fairness-verification} introduces fairness verification problem in NN. \secref{sec:approach} provides detailed description of the approach. \secref{sec:evaluation} describes the results and answers the research questions. Finally, 
%\secref{sec:disc} discusses future work in the area, 
% \secref{sec:threats} discusses the threats to validity, 
\secref{sec:related} describes the related work, \secref{sec:threats} discusses threats to validity, and \secref{sec:conc} concludes.

%%%%%%%% %%%%%%%% %%%%%%%% %%%%%%%% %%%%%%%% 
%%%%%%%% For the camera-ready version, the authors might want to:

% Done
% Soften the claim that Fairify is "the first technique to verify individual fairness of NN models in production." Replace with "Fairify operates on trained models while most of previous approaches operate during the training process."

% Clarify the choice of SMT and its limitations, as brought up by Reviewers B and C

% Done
% Discuss the group-vs-individual fairness distinction, in reference to Corbett-Davies and Goel

% Clarify the evaluation of partitioning and pruning, as raised in R3C1 in the response

% Address issues with scalability, as was done in the response (size of the dataset for industry vs. research standards) and highlight the need for future work

% Done
% Address the minor presentation issues raised by Reviewers A and B

% Consider renaming "WP" to "OP," as suggested by Reviewer C in the paper summary and as addressed in the response

%% file: preliminaries.tex
\section{Preliminaries}
\label{sec:background}
 
\textbf{Neural Networks (NN).} We consider NN as a directed acyclic graph (DAG), where the nodes hold numeric values that are computed using some functions, and edges are the data-flow relations. The nodes are grouped into layers: one input layer, one or more hidden layers, and one output layer.
More formally, a NN model $M: \mathbb{R}^n \rightarrow \mathbb{R}^m$ is a DAG with $k$ layers: $L_1, L_2, \dots , L_k$ where $L_1$ is the input layer and $L_k$ is the output layer. The size of each layer $L_i$ is denoted by $s_i$ and layer $L_i$ has the nodes $v_i^1, v_i^2, \dots , v_i^{s_i}$. Therefore, $s_1$ = $n$ (number of inputs) and $s_k$ = $m$ (number of output classes). In this paper, we consider the fully connected networks because of its success in real-world tasks \cite{paulsen2020reludiff,xiao2021self,zhang2020white}, where the value of a node is given by $v_i^j = \mathsf{NL}(\Sigma_t w_{i-1,t} \cdot v_{i-1}^t + b_i^j)$,
where a node $v_i^j$ is computed by the weighted sum $\mathsf{WS} = \Sigma_t w_{i-1,t} \cdot v_{i-1}^t + b_i^j$, and then applying a non-linear activation function $\mathsf{NL}$.
$\mathsf{WS}$ is computed from the neurons of preceding layer, associated weights of incoming edges $w_{i-1,t}$, and bias $b_i^j$. The weights and biases %\footnote{Bias of a network is different from the algorithmic bias/unfairness.} 
are constant real values in a trained NN, which are learned in the training phase. In practice, the ReLU activation is widely used as the NL function since it demonstrates good performance, and following the prior works in the area \cite{xiao2021self,katz2017reluplex, paulsen2020reludiff, urban2020perfectly}, we also considered ReLU based NN. ReLU is given by Eq. \eqref{eq:relu} which is a piecewise-linear ($\mathsf{PL}$) function.

% \begin{figure}[t]
% 	\centering
% 	\includegraphics[width=.9\columnwidth]{dnn.pdf}
% 	\caption{A fully-connected NN with ReLU activation}
% 	\label{fig:dnn}
% \end{figure}

{
\vspace{-10pt}
\small
\begin{align}
    \mathsf{ReLU}(x) = 
    \begin{cases}
        0 & \text{if } x \le 0 \text{ ; inactive neuron} \\
        x & \text{if } x > 0 \text{ ; active neuron}
    \end{cases}
    \label{eq:relu}
\end{align}
\vspace{-8pt}
}%

% \fignref{fig:dnn} depicts a neural network as a data flow graph. 
The neurons in the input layer accept data input values, which are passed to the following layers through the edges. Each neuron in the hidden and output layer computes its value by applying the weighted sum ($\mathsf{WS}$) function and then the $\mathsf{PL}$ function. 
The output neurons may have different $\mathsf{PL}$ functions (e.g., Sigmoid, Softmax) that computes the predictive classes in classification problem. 

\section{Fairness Verification}
\label{sec:fairness-verification}

In this section, we define the individual fairness verification problem and compare with closely related NN verifications.

\textbf{NN Verification.}
A property $\phi(M)$ of the NN model $M$ defines a set of input constraints as precondition $\phi_x$, and output constraints as postcondition $\phi_y$. Several safety properties of NN have been investigated in the literature \cite{nguyen2015deep,goodfellow2014explaining}. One of the most studied property is adversarial robustness \cite{huang2017safety, huang2015learning}. Given a model $M$ and input $x_0$, adversarial robustness ensures that a minimum perturbation to $x_0$ does not change the label predicted by the model, i.e., $\forall x'. \, ||x_0 - x'|| \le \delta \Rightarrow M(x_0) = M(x')$. The property ensures \textit{local} safety as it searches for a violation only in the neighborhood of $x_0$ with a distance $\delta$.
Different approaches have been proposed to guarantee safe region around the given input \cite{carlini2017towards, goodfellow2014explaining} or generate targeted attacks that violate the property \cite{szegedy2014google, nguyen2015deep, goodfellow2014explaining}. Unlike the robustness property, individual fairness requires \textit{global} safety guarantee.

\subsection{Individual Fairness}
\label{subsec:fairness-def}

The group fairness property considers average-case fairness, where some notion of parity is maintained between protected groups. Hence, this property fails to detect bias if same amount of discrimination is made between two groups \cite{galhotra2017fairness}. Individual fairness considers worst-case fairness, i.e., all similar input pairs get similar outcome \cite{galhotra2017fairness, john2020verifying}.
Our goal is to provide fairness guarantee which makes individual fairness property ideal for checking.
%Verifying individual fairness is different than 
%This notion of fairness is a global property as opposed to local robustness property where 

Suppose, $\mathcal{D}$ % = \{\mathsf{x}_1, \dots , \mathsf{x}_n\}$ 
is a dataset containing $n$ data instances, where each instance $x = (x_1, \dots , x_t)$ is a tuple with $t$ attribute values. 
The set of attributes is denoted by $A = \{A_1, \dots, A_t\}$ with its domain $I$, where $A_i = \{a \, | \, a \in I_i\}$. 
The set of protected attributes is denoted by $P$ where $P \subset A$. 
So, individual fairness is defined as follows, which is widely used in the literature \cite{zhang2020white,aggarwal2019black,udeshi2018automated,zheng2021neuronfair}.
%For binary classification problems $f: \mathbb{R}^n \rightarrow \{1, 0\}$. 
\begin{definition}
\label{def:fair1}
\textbf{(Individual Fairness)}
The model $M$ is individually fair if there is no such pair of data instances $(x, x')$ in the input domain such that:
\circled{1} $x_i = x'_i \text{ , } \forall i \in A \setminus P$, 
\circled{2} $x_j \neq x'_j \text{ , } \exists j \in P$, 
and \circled{3} $M(x) \neq M(x')$.
\end{definition}  

Intuitively, individual fairness ensures that any two individuals who have same attribute values except the protected attributes, get the same prediction. If there exists such pair, the property is violated and $(x, x')$ is considered as an individual bias
% \footnote{This terms \textit{bias} and \textit{unfairness} are used interchangeably in this paper.} 
instance. 
For example, suppose while predicting income of individuals ($>\$50$K or not), \textit{race} is considered as the protected attribute. \defref{def:fair1} suggests that if two individuals $(x, x')$ with different \textit{race} but exact same non-protected attributes, e.g., \textit{occupation}, \textit{age}, \textit{marital-status}, are predicted to the different class, then the model is unfair.
%  and $(x, x')$ is a bias instance.

The above definition is similar to the global robustness property introduced by \cite{katz2017reluplex, katz2017towards}. In global property checking, both the inputs $x$ and $x'$ can obtain any value within the domain and thereby is significantly harder to check. Whereas in local robustness an input is fixed and checking in the neighborhood is sufficient \cite{gopinath2018deepsafe, katz2017towards}. To check such global property, two copies of the same NN are encoded in the postcondition to check $M(x) \ne M(x')$. Hence, the approaches to verify local robustness properties (e.g., Reluplex) can not verify individual fairness \cite{katz2017towards, gopinath2018deepsafe}. \citeauthor{gopinath2018deepsafe} proposed a data-driven approach to assess global robustness \cite{gopinath2018deepsafe}. The approach requires labeled training data to cluster inputs and then casts the problem as local robustness checking. However, our goal is to verify individual fairness statically, i.e., we do not require training data or an oracle of correct labels.

In practice, definition \ref{def:fair1} can be a weaker constraint, and it can be \textit{relaxed} to ensure further fairness of the model. The attributes of $x$ and $x'$ are said to be relaxed when they are not equal. 
% Henceforth, the protected attribute is always relaxed in the above definition. 
The constraint \circled{1} above is relaxed on the non-protected attributes so that instead of equality, a small perturbation $\epsilon$ is allowed \cite{john2020verifying}.
For example, we can still consider $(x, x')$ as bias instance even if a non-protected attribute (e.g., \textit{age}) differs in a small amount (e.g., 5 years). 
Note that if a model is fair with respect to \defref{def:fair2}, it is also fair with respect to \defref{def:fair1} but the opposite is not true. Therefore, from the verification perspective, the relaxed query requires stronger certification and subsumes the basic fairness requirement.
% In our evaluation, we showed that Fairify can verify custom queries based on both the definitions.
\begin{definition}
    \label{def:fair2}
    \textbf{($\epsilon$-Fairness)}
    The model $M$ is individually fair if no two data instances $(x, x')$ in the input domain satisfy:
    \circled{1} $|x_i - x'_i | \le \epsilon_i \text{ , } \forall i \in A \setminus P$, \circled{2} $x_j \neq x'_j \text{ , } \exists j \in P$, and \circled{3} $M(x) \neq M(x')$, where $\epsilon$ is a small perturbation that limits the similarity. 
\end{definition}

For further practical benefit, we introduce the notion of targeted fairness, which imposes an arbitrary additional constraint on the inputs. For example, the developer might be interested in verifying whether the NN is fair in giving loans to individuals who have at least \textit{high-school} education, works more than \textit{50 hours-per-week}, and occupation is \textit{sales}. 

\begin{definition}
    \label{def:fair3}
    \textbf{(Targeted Fairness)}
    A target $T$ is a set of arbitrary linear constraints on the inputs of NN, i.e., $\{(l_i, u_i) \, | \, \forall i \in A, \, l_i \le x_i \le u_i\}$. Targeted fairness ensures that all valid inputs in the given target satisfy individual fairness.
\end{definition}

% The above definition is practical requirement for a specific classification task.
\citeauthor{holstein2019improving} identified a common difficulty in fair software development which is to diagnose and audit problems \cite{holstein2019improving}. 
The following developer's response \cite{holstein2019improving} conveys the utility of targeted verification and counterexamples, which was not  addressed in prior works.

{
\vspace{6pt}
% \small
\parbox{.92\columnwidth}{
    \textit{``If an oracle was able to tell me, `look, this is a severe problem and I can give you a hundred examples [of this problem],' [...] then it's much easier internally to get enough people to accept this and to solve it. So having a process which gives you more data points where you mess up [in this way] would be really helpful.''}
}
% \vspace{4pt}
}

%% file: approach.tex
\section{Approach}
\label{sec:approach}

\begin{figure*}[t]
	\centering
	\includegraphics[width=\linewidth]{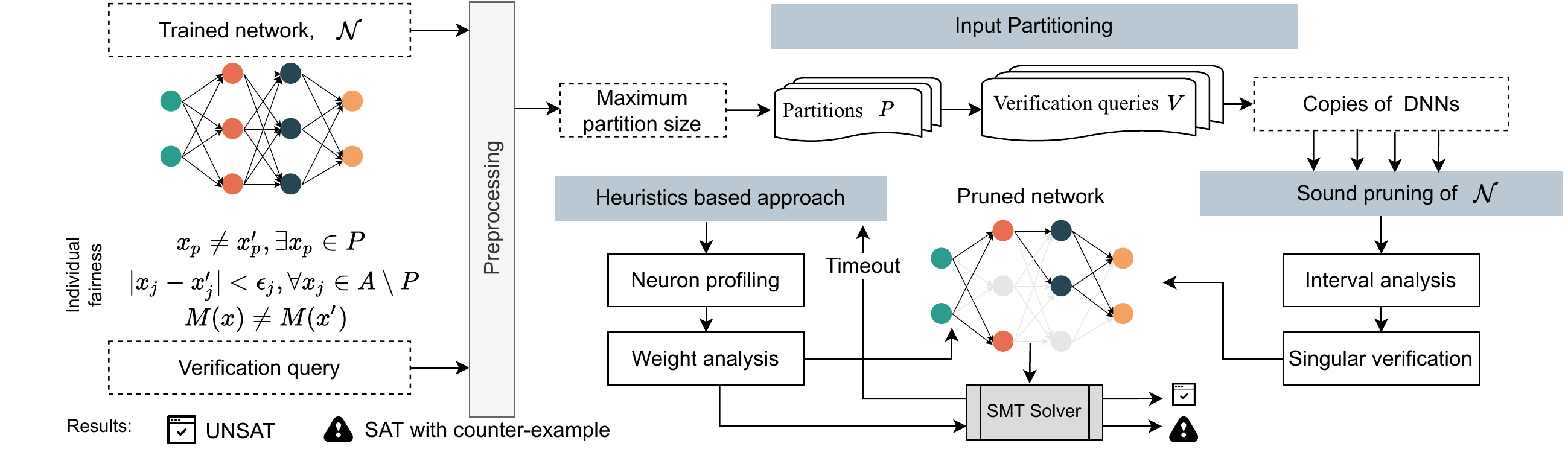}
	\caption{The overview of our approach for the fairness verification of neural networks}
	\label{fig:overview}
\end{figure*}

In this section, we formulate the fairness verification problem and describe our approach to verify the property. %The novelty in our approach is to reduce the problem into sub-problems which can be parallelized or run on multiple machines.

\subsection{Problem Formulation}
\label{subsec:problem}

The verification problem that we are interested in is, given the precondition on the input $x$ and the NN function $M$, the output $y = M(x)$ satisfies some postcondition. 
The preconditions and postconditions are designed to verify the fairness defined in \secref{sec:fairness-verification} on a trained model, where we have white-box access to the NN. In the definitions, \circled{1} and \circled{2} are the fairness preconditions, and \circled{3} is the fairness postcondition. Suppose, each input feature $x_i$ is bound by some lower bound $lb_i$ and upper bound $ub_i$ that is obtained from the domain knowledge. In this case, we obtain it from training data, e.g., while predicting income on Adult Census data, the \textit{work-hours} of individuals is $[1,100]$.
Let, $M$ be a classification network with one neuron in the output layer and suppose, Sigmoid function is applied on the output node $y$ to predict the classes. Sigmoid function is given by the equation: Sig$(y) = 1/(1+e^{-y})$. Therefore, for any two inputs $x, x'$, the postcondition that needs to be satisfied for fairness is: $M(x) = M(x')$. Here, we encode the negation of the postcondition in the verification query so that the solver provides counterexample with SAT when there is a violation.
Next, we compute the weakest precondition (WP) of the postcondition. A WP function gives the minimum requirements need to be satisfied to assert the postcondition. The WP computation is shown below, which transfers the fairness postcondition to the neurons of the output layer before applying the activation function, Sigmoid.

\vspace{-8pt}
{
\small
\begin{gather}
    WP(M(x) \neq M(x')) \equiv \text{Sig}(y) \neq \text{Sig}(y') \nonumber \\
    WP(\text{Sig}(y) \neq \text{Sig}(y')) \equiv (y < 0 \land y' > 0) \lor (y > 0 \land y' < 0) \nonumber
\end{gather}
}
\vspace{-8pt}

Similarly, for other non-linear activation functions, we can compute the WP. Another WP transfer used in our evaluation is Softmax function: $f(x_i) = e^{x_i} / \Sigma_j e^{x_j}$. The WP of Softmax for binary classification tasks is: $(y_0 > y_1 \Rightarrow y_0' < y_1') \land (y_0 < y_1 \Rightarrow y_0' > y_1')$.
Overall, this step makes the verification query simpler as well as the postcondition is reasoned on the network output layer. 
Thus, after reducing the postconditions, we have to verify the following constraint according to \defref{def:fair1}.

\vspace{-2pt}
{
\footnotesize
\begin{gather}
    \Big(
        \overbrace{
        \bigwedge_{\forall i \in A} lb_i \le x_i, x'_i \le ub_i}^\text{domain constraint, $\phi_x^d$}
    \Big) \land 
    \overbrace{
        \Big(
        \bigwedge_{\forall j \in A \setminus P} x_j = x'_j
    \Big) \land \nonumber
    \Big(
        \bigwedge_{\forall k \in P} x_k \neq x'_k
    \Big) 
    }^\text{fairness precondition, $\phi_x^f$} \land \\ \nonumber
    \underbrace{y = M(x), y' = M(x')}_\text{outputs} \Longrightarrow 
    \underbrace{\big(y < 0 \land y' > 0 \big) \lor \big(y > 0 \land y' < 0 \big)}_\text{fairness postcondition, $\phi_y$}
\end{gather}
}
\vspace{-2pt}

The above verification formula is defined using two copies of the same NN, its input constraints ($x$), and output constraints ($y$). The above constraint also support the \defref{def:fair2} and \defref{def:fair3}. For \defref{def:fair2}, we update the equality in $\phi_x^f$ with $|x_j - x'_j | \le \epsilon_j$. For \defref{def:fair3}, we use the bounds from target $T$ in $\phi_x^d$ instead of the original domains constraints.

Unlike local robustness \cite{katz2017towards, gopinath2018deepsafe}, the above fairness constraint requires twice as many input variables ($x_i$ and $x_i'$) and two copies of networks to obtain outputs ($y$ and $y'$). 
% Similar to \cite{urban2020perfectly}, in this paper we also considered fairness of NNs trained on structured or tabular data which has been of interest in many recent works \cite{zhang2020white,udeshi2018automated,albarghouthi2017fairsquare, bastani2019probabilistic,galhotra2017fairness}. 
% Since we do not consider fairness of image or natural language data, an off-the-shelve SMT solver can solve the constraint.
Other than that, the complexity of the constraint depends on the number of neurons and the structure of $M$. 
% We consider the formula as a verification query and pass that to SMT solver to get result. The difficulty in solving the query lies in the non-linear activation functions in the NN hidden layers Eq. \eqref{eq:relu}. The SMT solver has to verify the two phases of the non-linearity separately. 
In each neuron of the hidden layers, the solver divides the query into two branches: if the weighted sum is non-negative or else, as shown in Eq. \ref{eq:relu}. Thus, for $n$ neurons the query divides into $2^n$ branches, which can not be parallelized \cite{elboher2020abstraction}. 
In our approach, we identify the neurons that always remain inactive and remove them to reduce the complexity. Furthermore, we formulate the verification as an SMT based problem so that we can provide counterexample along the certification and verify $\epsilon$-fairness and targeted fairness. Existing abstract interpretation based techniques \cite{urban2020perfectly} can not be used towards that goal \cite{gopinath2019symbolic,gopinath2019property}.

\subsection{Solution Overview}
Fairify takes two inputs: trained neural network and fairness verification query. Then it performs three main steps: 1) input partitioning, 2) sound pruning, and 3) heuristic based pruning. An overview of the solution is depicted in \fignref{fig:overview}. Before going to step 1, Fairify preprocesses the verification query. It computes the WP of the postcondition to reduce the complexity of the verification formula.

After the preprocessing step, we have precondition defined on the neurons of input layer and postcondition defined on the neurons of the output layer. Then Fairify performs the input partitioning method. The main objective is two-fold: 1) the verification query becomes simpler, i.e., now the solver has to check less input region to prove the constraint; 2) given a smaller partition, the NN exhibits certain activation pattern so that we can prune the network. 
% Specifically, one of our key insights is that the NN trained on structured data exhibits sparseness in activation pattern, when we consider a smaller region.

After the input partitioning, Fairify first attempts the sound pruning approach where we use the tightened input bounds for the partitions to compute bound for each neuron. Here, we use the white-box access to the network weights and biases. Then we perform another step of verification on each neuron to prove its activation. This process is applied layer-wise, only one layer at a time. Hence, the checking takes very little time and identifies additional neurons that are inactive. Then inactive neurons are removed from the NN to reduce its size. Removing neurons from the NN largely reduces the complexity of the verification.
Finally, Fairify leverages an SMT solver to solve the query on the pruned version of the network. Fairify takes a soft-timeout as input parameter. When the soft-timeout is reached without any result, Fairify takes the heuristic based pruning approach and attempts to solve the verification query.

In this pruning approach, Fairify uses the network heuristics, e.g., weight magnitudes and their distribution among the neurons. We conduct a simulation on the NN to profile the neurons. If a neuron is never activated, it is a candidate for removal. Then we compare the candidates' distribution of magnitude with that of non-candidates. Thus, we identify additional neurons that are inactive in the given partition. Since this step may prune neurons which are rarely activated, the pruned version can lose small accuracy. Fairify takes a conservative approach which result in very little to no loss of accuracy. The approach design also allows \textbf{partial verification} for a subset of the partitions. The goal of verification is to provide SAT/UNSAT for as many partitions as possible within given time budget. Next, we describe three main components of Fairify in detail.

\subsection{Input Partitioning}
\label{subsec:input-part}

Fairify takes the parameter $MS$, which is the maximum size of an attribute $A_i$. Based on $MS$, Fairify automatically partitions the input domain into $m$ regions following the \algoref{algo:partition}. We first divide each attribute into $ \left\lceil (ub - lb)/MS \right\rceil $ partitions, and then taking each partition from each attribute, we get the regions. For each region, we assign a copy of the NN so that different pruning can be applied for each region as well as the query processing can be parallelized.
The verification results for the partitions are accumulated as follows. 

\begin{itemize}[leftmargin=*]
    \item If one partition is SAT, the whole problem becomes SAT. The counterexample shows a violation for the fairness query. 
    % However, it would be desired to check other partitions so that as many counterexamples can be found, which localizes fairness violations and provide insights for repair.
    \item If one partition provides UNSAT, the whole problem is not necessarily UNSAT. However, the verification provides guarantee that there are no two inputs possible in the given partition that violates the property. This provides partial certification which has benefits in provable repair.
    \item To prove that the whole problem is UNSAT, all the partitions required to be UNSAT.
\end{itemize}

\setlength{\textfloatsep}{4pt}
\begin{algorithm}[t]
    \footnotesize 
    \caption{Input partitioning based on domain constraints}
    \label{algo:partition}
    \begin{algorithmic}[1]

    \Require{Attributes $A$, Input domain $\mathbb{I}$, Max-size of attribute $MS$, Query $Q$}
    \Ensure{Region set $R$}
    % \change{\Ensure{Region set $R$, activation pattern $P$}}

    \Procedure{Input_Partitioner}{$A$, $\mathbb{I}$, $MS$, $\mathcal{N}$}
    \For{each attribute $A_i \in A$}
    \State $PT_{A_i} = $ []
        \If{$|A_i| > MS$}
            \State $low = LB(\mathbb{I}_i)$ \algorithmiccomment{lower bound of attribute}
            \State $high = UP(\mathbb{I}_i)$ \algorithmiccomment{upper bound of attribute}
            \If{$Q$ contains $|x - x | \le \epsilon_i | x,x' \in A_i$} \algorithmiccomment{$A_i$ is relaxed}
                \State $part = [low, high]$ \algorithmiccomment{no partitioning}
            \Else
                \While{$low \leq high$}
                    \State $high = low + MS$
                    \State $part = [low, high]$
                    \State Add $part$ to $PT_{A_i}$
                \EndWhile
            \EndIf
        \EndIf
    \EndFor 

    \State $R = \{ (part) \, | \, part \in PT_{A_i}, \forall A_i \in A \}$
    % \State $P = \textsc{Sound_Prunning}(\mathcal{N}, W, B)$ \algorithmiccomment{Store activation pattern of each partition}
    \State \Return{$R$}
    \EndProcedure

    \end{algorithmic}
\end{algorithm}

% While partitioning, it is important to note that the counterexamples ($x$ and $x'$) cannot lie in the two partitions.
% In other words, all partitions being UNSAT, if two counterexamples lie in two different partitions, then we might not detect that violation. Therefore, we do not partition the attributes that are relaxed on the inputs. For example, in \defref{def:fair1}, the protected attribute is always relaxed so that two counterexamples can take any two different values. 
% In case of \defref{def:fair2}, we also ignore partitioning the attribute which is relaxed in addition to the $PA$. 
Finally, after the partitioning, we shuffle the partitions to check different parts of the regions in the given timeout. 
% In our evaluation, we did not parallelize the execution, and hence the shuffling allowed to get counterexamples from various regions of the input domain. 
% Prioritizing the input partitions to cover more partitions in a quicker time can be a potential future work.

\subsection{Sound Pruning}
\label{subsec:sound-prune}

This step receives a number of verification problems from input partitioning. Each problem is associated with a copy of NN and the query $\phi$. The query is updated from the original query by tightening the bound of each attribute, given the smaller region. Now, we attempt to prune the network by removing the neurons that do not impact the prediction of the current copy of NN.

\textbf{Static analysis and Pruning of NN.} 
Before deployment, a trained NN can be assessed for certain properties. We obtain the weights and biases of the network to analyze its behavior.
Especially, each neuron does not contribute to the decision. The value of certain neurons depends on the incoming values, associated weights and bias.
Because of the ReLU activation function, many nodes become inactive, i.e., gets a zero valuation whenever the $\mathsf{WS}$ is negative. 
% \hridesh{Maybe I missed the definition of WS.} 
In our approach, we analyze activation pattern to remove those neurons which are always inactive. We perform such analysis using specific bounds on the input values. 
For example, %in \fignref{fig:dnn}, 
if we can assert $v_3^3 = 0$ for certain bounds on inputs $lb_1 < v_1^1 < ub_1$, $lb_2 < v_1^2 < ub_2$, then removing $v_3^3$ and the associated edges does not impact the value of $v_4^1$ and $v_4^2$.
After obtaining the weights and bias, we translate the NN into imperative program representation \cite{gopinath2019symbolic} so that it could be executed symbolically and constraint checker can assert first-order formulas. 
% Below is a Python program of canonical NN with one input-hidden-output layer. 
We leverage Numpy arrays and matrix operations to enable tracking the NN structure (e.g., layers, neurons) and perform network pruning.
% \begin{lstlisting}[language=Python]
% def net(x, w, b): # Inputs are Numpy arrays for input data, weight and bias
%     # Input layer
%     x1 = w[0].T @ x + b[0] # WS calculation using matrix multiplication
%     y1 = numpy.maximum(0, x1) # ReLU activation
%     # Hidden layer
%     x2 = w[1].T @ y1 + b[1]
%     y2 = numpy.maximum(0, x2) # ReLU activation
%     ... # Output layer
%     x_out = w[2].T @ y2 + b[2]
%     y = 1 / (1 + math.exp(-x_out)) # Sigmoid activation
%     return y # Output of NN 
% \end{lstlisting}
We take two steps to find such neurons: 1) interval analysis and 2) individual verification. 
% First, we compute bounds for each neuron in the hidden layers, and then perform layer-wise verification to prove that the given neuron is always inactive. 
The complete sound pruning algorithm is presented in \algoref{algo:sound}.

\setlength{\textfloatsep}{4pt}
\begin{algorithm}[t]
    %\small
    \footnotesize
    \caption{Sound pruning}
    \label{algo:sound}
    \begin{algorithmic}[1]

    \Require{Network $\mathcal{N}$, Weights $W$, Biases $B$, Region $R$}
    \Ensure{Pruned network $\mathcal{N}'$}

    \Procedure{Sound_Pruning}{$\mathcal{N}, W, B$}
    \State candidates = [] \algorithmiccomment{The candidate neurons for removal}
    \For{hidden neuron $v_i^j \ in$ Layer $L_i$}
        \State lb, ub = \textsc{Neuron_Bound}($v_i^j, W, B$)
        \If{ub $<$ 0}
            \State candidates.add($v_i^j$)
        \ElsIf{lb $>$ 0}
            \State $\mathcal{N}'$ = Update($\mathcal{N}$) \algorithmiccomment{Remove ReLU function from $v_i^j$}
        \Else
            \If{\textsc{Individual_Verification}($v_i^j, L_{i-1}$)}
                \State candidates.add($v_i^j$)
            \EndIf  
        \EndIf

    \EndFor
    \State $\mathcal{N}'$ = Update($\mathcal{N}$) \algorithmiccomment{Remove neurons in candidates}
    \State \Return{$\mathcal{N}'$}
    \EndProcedure

    \Procedure{Neuron_Bound}{$v_i^j, W, B$} \label{a:bound}
    \State $p_w$: non-negative incoming weights
    \State $n_w$: negative incoming weights
    \State $p_w = \{ w_{i-1, j} \, | \, w_{i-1, j} \ge 0, \forall j \in \{1, ..., |L_{i-1}|\}$ 
    \State  $n_w = \{ w_{i-1, j} \, | \, w_{i-1, j} \le 0, \forall j \in \{1, ..., |L_{i-1}|\}$ 
    
    \State $UB(v_i^j) = \Sigma_{w \in p_w} w * UB(v_{i-1}^j) + \Sigma_{w \in n_w} w * LB(v_{i-1}^j) + b_i^j$
    \State $LB(v_i^j) = \Sigma_{w \in p_w} w * LB(v_{i-1}^j) + \Sigma_{w \in n_w} w * UB(v_{i-1}^j) + b_i^j$
    
    \State \Return{$(UB(v_i^j), LB(v_i^j))$}
    \EndProcedure

    \Procedure{Individual_Verification}{$v_i^j, L_{i-1}$} \label{a:ind}
    \State precondition $pre = \{ LB \le v_{i-1}^j \le UB, \, \forall j \in \{1, ..., |L_{i-1}|\} \}$
    \State postcondition $post = v_i^j < 0$
    \State singular_verification = SMT-Solver($pre, post)$
    \If{singular_verification = UNSAT}
        \State \Return $true$
    \EndIf
    \State \Return $false$
    \EndProcedure

    \end{algorithmic}
\end{algorithm}

\subsubsection{Interval analysis}
We perform bound computation for the neurons in each hidden layer by using the bounds from its preceding layer, as shown in line-\ref{a:bound} in \algoref{algo:sound}. Here, we used interval arithmetic to compute the bounds. First, we separate the positive and negative incoming weights for a neuron $v_i^j$. Then we compute its maximum value by multiplying the upper bounds of the incoming neurons with positive weights and lower bounds with negative weights. We do the opposite to get the minimum value of $v_i^j$. 
The bounds are calculated without considering the activation functions but only the weighted sum. If the weighted sum is always $\leq 0$, we can remove the neuron from the NN with the edges connecting to it.
Since the interval arithmetic on multiplication does not lose any accuracy, the bounds always hold for the neurons. When we find that the upper bound of $v_i^j$ is $\le 0$ we mark $v_i^j$ for removal. The key intuition for this step is that since we have tightened bounds for the input layer (after partitioning), the bounds of hidden neurons are also tighter. Therefore, we can remove more neurons by applying this step on the partitioned regions. However, the bounds of many neurons, especially in deep layers, may not be tight enough to prove it is inactive. So, we apply the next step of individual verification on each neuron.

\subsubsection{Individual verification}
We formulate verification query for each hidden neuron to further prove whether it is inactive. The precondition for the verification query for neuron $v_i^j$ consists of the bounds $v_{i-1}^j$, and the postcondition is $v_i^j < 0$. Then we leverage the SMT solver to prove that the neuron is inactive. We run these individual verification queries on neurons in the increasing order of the layers. So, all the neurons of one layer are verified before going to the next layer. Similar to the bound analysis, the verification query does not include any non-linear activation functions i.e., ReLU. In addition, the query is designed with the precondition that contains values of only the immediate preceding layer. 
Furthermore, we apply individual verification query only on the neurons that are not already pruned by interval analysis.
Therefore, the individual verification queries are faster and can find more inactive neurons that were not found in the previous step.

\subsection{Heuristic Based Pruning}
\label{subsec:heuristic-prune}

We investigated the real-world models used in the fairness problems on structured data, and we found that the datasets are far sparse from problem that involve high dimensional data such as image classification or natural language processing. For example, each instance in the MNIST dataset is (28, 28) Numpy array, whereas a popular Adult Census dataset contains data instances each with the shape (1, 13). It causes the trained NN to have many neurons that are never activated or only activated for a specific region. Applying the methods in the previous stage, we could detect some provably guaranteed inactive neurons in the NN. But there can be further such neurons which are not activated for the given region in practice. How can we detect those neurons?

A variety of network heuristics has been studied for different purposes such as quantization, efficiency, accuracy, etc. \cite{dong2017learning, guo2016dynamic, han2015learning}, which can be performed in different stages of the development e.g., before or during training. Here, we proposed a novel conservative approach of NN pruning for fairness verification based on heuristics. Our goal is to reduce the network size without losing accuracy to the extent possible.
%Magnitude based scoring showed success in the past to identify activation patterns \cite{gale2019state, han2015learning}. 

The algorithm for heuristic based pruning is shown in \algoref{algo:heuristic}. For learning the heuristics of the network, we run the neuron profiler. The objective is to get the distribution of magnitudes of each neuron. First, we create a simulated dataset $\mathcal{D}$ of size $S$ by uniformly generating valid inputs from attribute domain. Then the network $M$ is run $S$ times to record the values for each neuron. In practice, we used $S = 1000$, which allows to separate most of the candidates and non-candidates. Initially, all the neurons are selected as candidates for removal. If a neuron gets non-zero value for at least one execution of the simulation, we remove that from the candidate list. Thus, after recording the values of both the candidates and non-candidates, we confirm that whether the distribution of positive magnitude differs significantly. The intuition is that even if the neuron in the candidate is inactive for all the executions, it might get activated for some valid input data. However, the magnitude distribution of the neurons suggests whether it is active or not. Our evaluation shows that there is a significant difference in their distribution. For example, we observed that in the first hidden layer of the NNs, the mean and median of the positive magnitude of non-candidates is at least 10 times larger.

\setlength{\textfloatsep}{3pt}
\begin{algorithm}[t]
    \footnotesize
    %\small
    \caption{Heuristics based pruning}
    % \hspace{\algorithmicindent} \textbf{Input} \\
    % \hspace{\algorithmicindent} \textbf{Output} 
    \label{algo:heuristic}
    \begin{algorithmic}[1]

    \Require{Network $\mathcal{N}$, Domain $\mathbb{I}$, Simulation size $S$, Tolerance $T$}
    \Ensure{Pruned network $\mathcal{N}'$}

    \Procedure{Heuristic-Prune}{$\mathcal{N}$, $\mathbb{I}$, $S$}
    \State $candidates, Mag^+, Mag^-$ = \textsc{Neuron-Profiler}($\mathcal{N}, S$)
    \State $Dist_{candidate}$ \algorithmiccomment{Distribution of the candidates} 
    \State $Dist_{noncandidate}$ \algorithmiccomment{Distribution of the non-candidates}

    \For{$v_i^j$ in hidden layer $L_i$}
        \State check candidates and non-candidates distribution differs
        \If{$Mag^+(v_i^j) < T\text{-level of } Dist_{noncandidate} $}
            \State Remove $v_i^j$ from $M$
        \EndIf
    \EndFor 
\EndProcedure

\Procedure{Neuron-Profiler}{$\mathbb{I}$, $S$}
    \State $\mathcal{D} = \{X | X_i \in \mathbb{I} \text{ and } |X| = S\}$
    \State $candidates = \{ v_i^j \, | \, v_i^j \in L \}$ \algorithmiccomment{L is a hidden layer in $M$}
    
    \For{$X_i \in \mathcal{D}$}
        \State Run $\mathcal{N}(X_I)$ to record heuristic
        \For{hidden neurons $v_i^j$ in $\mathcal{N}$}
            \If{$v_i^j$ is active}
                \State $candidates.remove(v_i^j)$
            \EndIf
            \If{$v_i^j \ge 0$} \algorithmiccomment{Positive magnitude}
                \State $Mag^+ = |v_i^j|$
            \Else \algorithmiccomment{Negative magnitude}
                \State $Mag^- = |v_i^j|$
            \EndIf
        \EndFor
    \EndFor
    \State \Return{$candidates, Mag^+, Mag^-$}
\EndProcedure

\end{algorithmic}
% \vspace{-5mm}
\end{algorithm}

The above heuristic allows to compare the magnitude of each candidate neuron and conservatively select it for removal. Fairify takes a tolerance level as input for this comparison. In our evaluation, we used 5\% as the tolerance level. Fairify selects the candidates which has a magnitude less than the 5-percentile of the non-candidate neurons. The weight magnitude difference is more prominent in deep layers than the shallow ones. So, we do layer-wise comparison and candidate selection based on the heuristics. 

One might argue that since this pruning does not provably guarantee the same outcome as the original NN, the verification result can be inaccurate. However, the idea here is to use the pruned version of the NN in production as opposed to the original NN. If we certify the pruned NN and observe little or no accuracy decrease, then the pruned and verified model itself can be used in the production. Thus, we can preserve the sanity of the verification results. Our evaluation shows that it is possible to carefully select the heuristic so that NN is pruned conservatively, which affects accuracy negligibly but enables faster verification.

%% file: evaluation.tex
\section{Evaluation}
\label{sec:evaluation}

In this section, first, we discuss the experimental details and then answer three research questions regarding the utility, scalability, and performance of our approach. 

\subsection{Experiment}

\subsubsection{Benchmarks}

The verification benchmark is a crucial part of the evaluation. 
We undertook several design considerations to enable fairness verification in development pipeline. Therefore, unlike prior works we created comprehensive real-world benchmark of NN models from theory and practice.

We evaluated Fairify on three popular fairness datasets i.e., Bank Marketing (BM), Adult Census (AC), and German Credit (GC) \cite{zhang2020white,galhotra2017fairness,aggarwal2019black,biswas21fair}. The benchmark models (\tabref{tab:benchmark}) are collected from four different sources. 
First, we followed the methodology of \citeauthor{biswas20machine} to collect real-world NNs from Kaggle \cite{biswas20machine}. We searched all the notebooks under the three datasets in Kaggle and found 16 different NNs.
Second, \citeauthor{zhang2020white} used 3 NNs trained on the aforementioned datasets \cite{zhang2020white}.
Third, \citeauthor{udeshi2018automated} \cite{udeshi2018automated} evaluated fairness testing on one NN architecture, which is further used by \citeauthor{aggarwal2019black} \cite{aggarwal2019black} on the three datasets. We found that two of these three models are also implemented in Kaggle notebooks.
% Fourth, FairSquare used 3 NN models in their evaluation \cite{albarghouthi2017fairsquare}.
Finally, NN models AC8-12 were created by \cite{urban2020perfectly,mazzucato2021reduced} for dependency fairness certification.
% \citeauthor{mazzucato2021reduced} further used those models to improve performance of Libra \cite{mazzucato2021reduced}.
% Although Fairify provides different output than Libra, we showed the verification results in \tabref{tab:libra}.
Thus, we created a fairness benchmark of 25 NNs.
% The models have 3 to 11 layers with 10 to 221 neurons. 
The networks used for these fairness problems are fully connected with ReLU activation functions, which is also observed by many prior works \cite{zhang2020white,albarghouthi2017fairsquare,urban2020perfectly,mazzucato2021reduced}. 
% In addition, the accuracies of the models are very high compared to the other ML classifiers trained on these datasets.
The models and datasets are placed into our replication package to make the tool self-contained. The details of the datasets are as follows:

\textit{Bank Marketing} dataset contains marketing data of a Portuguese bank which is used to classify whether a client will subscribe to the term deposit \cite{moro2014data}. It has 45,000 data instances with 16 attributes.
%  and \textit{age} is considered protected.
\textit{German Credit} dataset contains 1000 data instances of individuals with 20 attributes who take credit from a bank \cite{germanuci}. The task is to classify the credit risk of a person.
% , whereas \textit{sex} and \textit{age} are considered protected. 
\textit{Adult Census} dataset contains United States census data of 32,561 individuals with 13 attributes \cite{kohavi1996scaling}. The task is to predict whether the person earns over \$50,000.
%  Here, \textit{race} and \textit{sex} are considered as the protected attributes.

\input{benchmark.tex}

\subsubsection{Experiment setup}
Fairify is implemented in Python and the models are trained using Keras APIs \cite{urban2020perfectly}. Following the prior works in the area \cite{albarghouthi2017fairsquare}, we used Z3 \cite{de2008z3} as the off-the-shelf SMT solver for manipulating the first-order formulas. However, other SMT solvers can also be used, since our technique of input partitioning and network pruning can work independent of any SMT solver.
% In addition, Fairify is designed in such a way that the developers can run it on personal computers. 
The experiments are executed on a 4.2 GHz Quad-Core Intel Core i7 processor with 32 GB memory.

\textit{\textbf{Input.}} Fairify takes the trained NN model, input domain, the verification query, maximum partition size, and timeout as inputs. The trained models are saved as \texttt{h5} files, from which Fairify extracts the necessary information e.g., weight, bias, structure. 
% The domain information includes the lower and upper bound of each input attributes in the dataset. 
The query includes the name of the protected attributes and relaxation information of the other attributes.

\textit{\textbf{Output.}} Fairify provides verification result for each partition. The results include verification (SAT/UNSAT/UNKNOWN), counterexample (if SAT), and pruned NN for the partitions.

\subsection{Results}
We answered three broad research questions for evaluation:
\begin{itemize}
  \item RQ1: What is the \textbf{utility} of our approach?
  \item RQ2: Is Fairify \textbf{scalable} to relaxed fairness queries?
  \item RQ3: What is the \textbf{performance} with respect to time and accuracy of the approach?
\end{itemize}

\subsubsection{Utility} 
First, we evaluated the baseline fairness verification results and then compared with our approach, which is shown in \tabref{tab:utility}.
Since GC and AC contained multiple protected attributes (PA), we setup multiple verification for each of those NNs. However, the verification results are consistent over different PAs. Hence, we showed result for one PA for each dataset in \tabref{tab:utility}. The whole result is also available in our supplementary material \cite{supplement}. 
In this RQ, we evaluated the fairness defined in Eq. \eqref{def:fair1}.
% , which requires two individuals with same attributes (except the protected attribute) are classified to the same class. 
We discussed the verification of relaxed fairness constraints Eq. \eqref{def:fair2} and \eqref{def:fair3} in RQ2. 

For the baseline verification, we encoded fairness property into satisfiability constraints, and then passed the \textit{original NN} and constraints to the SMT solver. In addition, we attempted to verify the original NN with input partitioning. On the other hand, Fairify used the method of input partitioning and NN pruning to demonstrate the improvement over the baseline. 
For the baseline, we set a timeout of 30 hours for the solver and run the verification for the models in our benchmark. 
% The results are shown in the first two columns in \tabref{tab:utility}. 
Only one model (AC6) could be verified within the timeout.
% The solver returned SAT with counterexample for AC6. 
For the other models, the solver reported UNK i.e., the model could not be verified within the time limit.
Furthermore, we try the baseline verification with input partitioning and run the queries on the NNs but we get the same verification result as the baseline.
The main takeaway is that the difficulty in verifying fairness lies in the complex structure of the NNs. Only input partitioning reduces the input space to be verified, but that does not reduce the complexity of NN. Fairify combines partitioning and pruning to enable network complexity reduction, which made the verification feasible.
%The baseline models fails or keep running because each ReLU node divides the query into two parts. With the number of ReLU nodes the complexity grows exponentially. Our approach removes the redundant neurons to make the verification tractable. 

\input{utility.tex}

\textit{\textbf{How effective is our approach to verify fairness of NN?}} 
We presented the verification results of 25 NNs in \tabref{tab:utility}. We found that Fairify produces verification results very quickly compared to the baseline. For each model, we run the verification task for 30 minutes (hard-timeout). Since we have divided the single verification into multiple partitions, we set 100 seconds as the soft-timeout for the SMT solver, which means that the solver gets at most 100 seconds to verify. When the result of the sound verification is UNK, then Fairify attempts the heuristic based pruning and runs the SMT solver for another 100 seconds. The goal of using a short soft-timeout is to show the effectiveness of our approach over baseline.

The results show that 19 out of 25 models were verified within the 30 minutes timeout. The models that could not be verified in that time period were considered again with scaled experiment setup in RQ2.
Fairify takes the maximum size of an attribute (MS) as an input to automatically partition the input region. The user can select MS based on the range of the attributes in the dataset. The timeout and MS can be configured based on the budget of the user. For BM and GC models, we used 100 as MS, and for AC models we used 10 as MS. To that end, Fairify divided input region into 510, 201, and 16000 partitions using \algoref{algo:partition}. Here is an example verification task from our evaluation:

\textbf{Example:}
While verifying AC3 (\textit{race} as PA), Fairify takes the following partition as a sub-problem. First, it attempts sound pruning and achieves 86.27\% compression. Then it runs verification query for 100 seconds and reports SAT with the counterexamples \textbf{C1} and \textbf{C2} in 21.47 seconds. Note that, these are two inputs for which the NN is not fair. Here, the two individuals had the same attributes except for \textit{race} but were classified as \textit{bad} and \textit{good} credit-class, respectively.

\vspace{5pt}
{
\footnotesize
\noindent\fbox{%
\parbox{.97\columnwidth}{
    {workclass: [0, 6], marital-status: [0, 6], relationship: [0, 5], race: [0, 4], sex: [0, 1], age: [80, 89], education: [0, 9], education-num: [11, 16], occupation: [0, 9], capital-gain: [10, 19], capital-loss: [0, 9], hours-per-week: [51, 60], native-country: [30, 39]}
    % \vspace{-5pt}
\begin{center}
    \textbf{C1:} $[89, 6, 9, 14, 1, 0, 0, 0, \textcolor{red}{1}, 14, 8, 59, 39]$
    
    \textbf{C2:} $[89, 6, 9, 14, 1, 0, 0, 0, \textcolor{red}{0}, 14, 8, 59, 39]$
\end{center}
\vspace{-4pt}
}}
% \vspace{-2pt}
}

% Similar to the above example, Fairify checks a number of partitions for each model and reports the results.
Depending on the complexity of each NN, Fairify could complete verification for a certain number of partitions in the given timeout period. Whenever we get SAT for at least one partition, the whole verification is SAT. Fairify also reports the counterexample when it reports SAT. We included the detailed results for each partition including all generated counterexamples in our replication package \cite{replication}.

\textit{\textbf{Is the NN pruning effective for verifying fairness?}}
The baseline models cannot be verified in a tractable amount of time. After being able to prune the models significantly, we could verify them in a short period. We computed the amount of pruning applied to the models. For the original $M$ and pruned version $M'$, compression ratio is calculated using the following formula: $1 - |M'| / |M|$, where $|M|$ is the number of neurons in $M$. The average compression percentage in \tabref{tab:utility} shows that Fairify could reduce the size of NN highly in all the models.
Heuristic based pruning is only applied when Fairify cannot get verification result within the soft-timeout. Furthermore, heuristic based pruning is applied on the already pruned version of the NN. Therefore, 
% for our design choice of adopting a conservative approach mentioned in \secref{subsec:heuristic-prune}, 
the compression ratio for heuristics based pruning is lower.
We found that 13.98\% times Fairify attempted heuristic pruning, 19.38\% of those times Fairify provided SAT or UNSAT result, meaning that the additional pruning of Fairify helped to complete the verification.
In the example showed above, Fairify did not attempt a heuristic based pruning, since it got the result in sound pruning step. 
% Therefore, there is no loss of accuracy. 
% For further validation purpose, we run the original models on the 1,302 reported counterexamples and observe that those violate the fairness property of interest.

\textit{\textbf{Can Fairify be used to localize fairness defects?}}
For ML models, it is difficult to reason or find defects since the model learns from data. In fairness problems, oftentimes the model learns from biased data or augments the bias during training. If we can filter the input domain where the model is unfair, then it would guide fairness repair. For example, in model BM3, Fairify provides SAT for 27 partitions and UNSAT for 108 partitions. The developer can leverage the verification result to further improve the training data and retrain the network. Another novelty is that since Fairify verifies multiple copies of pruned NN, the developer may choose to deploy those pruned versions into production. When an input comes to the system, the software can choose to use the verified copies of NN.
% Thus, the system can guarantee fairness property by choosing not to use defective copies of the model.
Qualitatively, Fairify provides the following two main utilities:

\paragraph{Tractability and speedup} The results showed that the verification for the NN becomes tractable when our approach is applied.
Furthermore, the partition size and timeout can be tuned based on the complexity of the NN. In our evaluation, we used a short timeout of 30 minutes \cite{paulsen2020reludiff}. The verification has been possible for most of the models in this time because of successful NN pruning. In addition, as soon as the first SAT is found, the developer may choose to stop running verification.

\paragraph{Partial verification} Even after getting SAT for one partition, our evaluation continued verification for other partitions, which essentially provides partial verification. For example, partitions with UNSAT imply that the NN is fair for that specific input region. Similarly, SAT with the counterexamples for a partition can be used towards repairing the NN, which is a potential future work.

\subsubsection{Scalability}
\label{subsec:salability}

In RQ1, we have set a small timeout and could verify 19 out of 25 models. In this RQ, we set a scaled experiment setup to further verify the remaining 6 models. We noticed that these 6 NNs are complex because of more number of layers and neurons.
This time we used a hard-timeout of 1 hour and soft-timeout of 200 seconds, essentially doubling the timeouts. We also reduced the maximum size of the partition (MS). The results are shown in \tabref{tab:stress}. We have verified the 3 out of the 6 models within that 1 hour. The results demonstrate the scalability of our approach for more complex NNs.
We found that for some partitions, the compression ratio is more, and hence the SMT solver could verify quickly. So, it would be an interesting future work to prioritize the partitions to verify for efficiency. Next, we showed whether our approach can verify complex verification queries for all the models.

\input{stress.tex}

\input{relaxed.tex}

\textit{\textbf{Can Fairify verify relaxed and targeted verification queries?}}
To answer this question, we created relaxed and targeted fairness queries according to \defref{def:fair2} and \defref{def:fair3} respectively. 
The verification results for the relaxed queries are presented in \tabref{tab:relaxed}.
First, for the \textbf{relaxation} of individual fairness, we define small perturbation ($\epsilon$) on the non-protected attributes so that two individuals are considered similar even if they are not equal in any non-protected attributes. Those two individuals still have to be classified to the same class to ensure fairness. Therefore, the relaxation on the queries impose stricter fairness requirement. We created six different such queries run Fairify on all the models. In each query, we selected an additional non-protected attribute, which was relaxed. For example, a verification query for AC is ($\phi_{r31}$): \textit{Is the NN fair with respect to race, where any two people are similar irrespective of their marital status?}
We found that 21/25 for $\phi_{r1}$ and 22/25 for $\phi_{r2}$ were verified within one hour. Compared to \tabref{tab:utility}, the number of SAT found is more since $\epsilon$-fairness is a stricter requirement and creates possibility of more counterexamples.
Second, for the \textbf{targeted} verification, we created six other fairness queries that targets a specific population. For example, $\phi_{t32}$ verifies that \textit{whether the NN is fair for people who have bachelor or doctorate education}. 
The results show that Fairify can verify most of the models in a quick time. 
Similar to the scaled experiment setup in \tabref{tab:stress}, we used a smaller partition size in this experiment. 
Although the queries are more complex, Fairify could provide counterexample or certification for many partitions. 
% The number of partitions that are verified depends on the relaxed attribute and the size of the NN.

\subsubsection{Performance}

\textit{\textbf{How quickly the verification is done and what is the overhead?}}
Depending on the verification query and the model, the verification time varies. \tabref{tab:utility} presents the average time taken for the partitions of each model. We showed the time taken by the SMT solver to output a result in the sound verification phase, and in heuristic verification phase. 
We also calculated the total time taken to complete verification, which includes the partitioning and pruning discussed in \secref{sec:approach}. The results show that the additional time taken for the three steps in our approach is negligible with respect to the time taken by the SMT solver. 
The partitioning of inputs is a one time step. For each partition, we apply pruning once or twice. However, pruning is static operation without any complex constraint solving. The only part of our pruning steps that take more time compared to other steps is the individual verification of each neuron which uses the SMT solver. But in that verification, only one layer is considered at a time, activation functions are excluded, and the number of constraints is at most the number of hidden neurons in a layer. Therefore, 
excluding the time taken by the SMT solvers to solve the final constraints, 
Fairify did not take more than 10 seconds for any model in that step.

\textit{\textbf{What is the performance of Fairify compared to the related work?}}
As described in \secref{sec:fairness-verification}, individual fairness verification of NN can not be done with existing robustness checkers \cite{gopinath2018deepsafe, katz2017towards}. \citeauthor{shriver2021dnnv} proposed a framework called DNNV \cite{shriver2021dnnv} that incorporates the state-of-the-art NN verifiers, e.g., ReluPlex, Marabou, etc. We tried verifying the fairness queries using those DNNV verifiers, however, they can verify queries with a single network input variable only \cite{shriver2021dnnv, dnnv}. 
% Hence, it can not be used for global property verification, e.g, individual fairness.
% Although we could verify local robustness properties for the models in our benchmark in DNNV, the fairness queries result in an error -- \textit{exactly one network input is required}.
Other than that, we verified the models (AC8-12) from Libra \cite{urban2020perfectly,mazzucato2021reduced} shown in \tabref{tab:utility} and \tabref{tab:relaxed}.
% The models contained 10, 12, 20, 40, 45 neurons with 2, 4, 4, 4, 9 hidden layers respectively. 
Fairify could verify the models for all the queries except AC11-12 for two out of four queries. The overall coverage is less than that reported by \cite{mazzucato2021reduced} because Fairify verifies a different property, the configuration is lower, and experiment setup is different. 
% Fairify divides input domain based on the configuration and verifies each partition but 
Libra computes abstract domain and projects into the input space to find biased region. So, the precision of Libra depends on the chosen abstract domain. On the other hand, Fairify can be configured for arbitrary queries, partial verification, and additionally we provide counterexample and pruned NN as output.
% Furthermore, Fairify can localize verification result through the customization of targeted region of interest and relaxation of queries. 
Therefore, Fairify can be more appropriate for defect localization or repair. One limitation of Fairify is that when the NN is deep and wide at the same time, the pruning ratio is less, and the SMT solver may return UNK in the given timeout. However, developers can configure less conservative approach in heuristic pruning to circumvent the problem. Thus, dynamical tuning of configuration could be a potential future work for Fairify.

\textit{\textbf{What is the accuracy loss of heuristic-based pruning?}}
We calculated the accuracy of the pruned NNs for each partition. When no heuristic based pruning is done, there is no accuracy loss. However, when Fairify applies heuristic based pruning there might be accuracy loss compared to the original NN. 
We took a conservative approach for the heuristic based pruning. Therefore, there was no accuracy loss for the pruned NN. 
Note that even if there is a small accuracy reduction for a heuristic based approach, the developer may choose to deploy the pruned version of the NN as opposed to the original one.

%% file: benchmark.tex
\begin{table}[t]
      % \vspace{2mm}
      \centering
      % \scriptsize
      % \footnotesize
      \setlength\tabcolsep{5.5pt}
      \caption{The NN benchmark for fairness verification}
      \begin{tabular}{|l|llrrr|}
      \hline
      \rowcolor[rgb]{ .851,  .851,  .851} \textbf{Dataset} & \textbf{Model} & \textbf{Source} & \multicolumn{1}{l}{\textbf{\#Layers}} & \multicolumn{1}{l}{\textbf{\#Neurons}} & \multicolumn{1}{l|}{\textbf{Acc \%}} \\
      \hline
            \multicolumn{1}{|l|}{\multirow{8}[2]{*}{\begin{sideways}Bank Marketing\end{sideways}}} & BM1   & Kaggle & 4     & 97    & 89.20 \\
            & \cellcolor[rgb]{ .949,  .949,  .949}BM2 & \cellcolor[rgb]{ .949,  .949,  .949}Kaggle & \cellcolor[rgb]{ .949,  .949,  .949}4 & \cellcolor[rgb]{ .949,  .949,  .949}65 & \cellcolor[rgb]{ .949,  .949,  .949}88.76 \\
            & BM3   & Kaggle, \cite{aggarwal2019black,udeshi2018automated} & 3     & 117   & 88.22 \\
            & \cellcolor[rgb]{ .949,  .949,  .949}BM4 & \cellcolor[rgb]{ .949,  .949,  .949}Kaggle & \cellcolor[rgb]{ .949,  .949,  .949}5 & \cellcolor[rgb]{ .949,  .949,  .949}318 & \cellcolor[rgb]{ .949,  .949,  .949}89.55 \\
            & BM5   & Kaggle & 4     & 49    & 88.90 \\
            & \cellcolor[rgb]{ .949,  .949,  .949}BM6 & \cellcolor[rgb]{ .949,  .949,  .949}Kaggle & \cellcolor[rgb]{ .949,  .949,  .949}4 & \cellcolor[rgb]{ .949,  .949,  .949}35 & \cellcolor[rgb]{ .949,  .949,  .949}88.94 \\
            & BM7   & Kaggle & 4     & 145   & 88.70 \\
            & \cellcolor[rgb]{ .949,  .949,  .949}BM8 & \cellcolor[rgb]{ .949,  .949,  .949}\cite{zhang2020white} & \cellcolor[rgb]{ .949,  .949,  .949}7 & \cellcolor[rgb]{ .949,  .949,  .949}141 & \cellcolor[rgb]{ .949,  .949,  .949}89.20 \\
      \hline
      \multicolumn{1}{|l|}{\multirow{5}[2]{*}{\begin{sideways} {\makecell{German\\Credit}}\end{sideways}}} & GC1   & Kaggle & 3     & 64    & 72.67 \\
            & \cellcolor[rgb]{ .949,  .949,  .949}GC2 & \cellcolor[rgb]{ .949,  .949,  .949}\cite{aggarwal2019black,udeshi2018automated} & \cellcolor[rgb]{ .949,  .949,  .949}3 & \cellcolor[rgb]{ .949,  .949,  .949}114 & \cellcolor[rgb]{ .949,  .949,  .949}74.67 \\
            & GC3   & Kaggle & 3     & 23    & 75.33 \\
            & \cellcolor[rgb]{ .949,  .949,  .949}GC4 & \cellcolor[rgb]{ .949,  .949,  .949}Kaggle & \cellcolor[rgb]{ .949,  .949,  .949}4 & \cellcolor[rgb]{ .949,  .949,  .949}24 & \cellcolor[rgb]{ .949,  .949,  .949}70.67 \\
            & GC5   & \cite{zhang2020white}    & 7     & 138   & 69.33 \\
      \hline
      \multicolumn{1}{|l|}{\multirow{12}[2]{*}{\begin{sideways}Adult Census\end{sideways}}} & \cellcolor[rgb]{ .949,  .949,  .949}AC1 & \cellcolor[rgb]{ .949,  .949,  .949}Kaggle & \cellcolor[rgb]{ .949,  .949,  .949}4 & \cellcolor[rgb]{ .949,  .949,  .949}45 & \cellcolor[rgb]{ .949,  .949,  .949}85.24 \\
            & AC2   & Kaggle, \cite{aggarwal2019black,udeshi2018automated} & 3     & 121   & 84.70 \\
            & \cellcolor[rgb]{ .949,  .949,  .949}AC3 & \cellcolor[rgb]{ .949,  .949,  .949}Kaggle & \cellcolor[rgb]{ .949,  .949,  .949}3 & \cellcolor[rgb]{ .949,  .949,  .949}71 & \cellcolor[rgb]{ .949,  .949,  .949}84.52 \\
            & AC4   & Kaggle    & 4     & 221   & 84.86 \\
            & \cellcolor[rgb]{ .949,  .949,  .949}AC5 & \cellcolor[rgb]{ .949,  .949,  .949}Kaggle & \cellcolor[rgb]{ .949,  .949,  .949}4 & \cellcolor[rgb]{ .949,  .949,  .949}149 & \cellcolor[rgb]{ .949,  .949,  .949}85.19 \\
            & AC6   & Kaggle & 4     & 45    & 84.77 \\
            & \cellcolor[rgb]{ .949,  .949,  .949}AC7 & \cellcolor[rgb]{ .949,  .949,  .949}\cite{zhang2020white} & \cellcolor[rgb]{ .949,  .949,  .949}7 & \cellcolor[rgb]{ .949,  .949,  .949}145 & \cellcolor[rgb]{ .949,  .949,  .949}84.85 \\
            & AC8   & \cite{urban2020perfectly,mazzucato2021reduced}    & 4     & 10    & 82.15 \\
            & \cellcolor[rgb]{ .949,  .949,  .949}AC9 & \cellcolor[rgb]{ .949,  .949,  .949}\cite{urban2020perfectly,mazzucato2021reduced} & \cellcolor[rgb]{ .949,  .949,  .949}6 & \cellcolor[rgb]{ .949,  .949,  .949}12 & \cellcolor[rgb]{ .949,  .949,  .949}81.22 \\
            & AC10  & \cite{urban2020perfectly,mazzucato2021reduced}    & 6     & 20    & 78.56 \\
            & \cellcolor[rgb]{ .949,  .949,  .949}AC11 & \cellcolor[rgb]{ .949,  .949,  .949}\cite{urban2020perfectly,mazzucato2021reduced} & \cellcolor[rgb]{ .949,  .949,  .949}6 & \cellcolor[rgb]{ .949,  .949,  .949}40 & \cellcolor[rgb]{ .949,  .949,  .949}79.25 \\
            & AC12  & \cite{urban2020perfectly,mazzucato2021reduced}    & 11    & 45    & 81.46 \\
            \hline
      \end{tabular}%
      \label{tab:benchmark}%
      % \vspace{-8pt}
      \end{table}%

%% file: utility.tex
\begin{table}[t]
      %\vspace{2mm}
      \centering
      % \footnotesize
      \scriptsize
      \setlength\tabcolsep{1.5pt}
      \caption{Fairness verification results for NN}
        \begin{tabular}{|c|c|c|rrrrrrrrrrrr|}
        \hline
        \rowcolor[rgb]{ .851,  .851,  .851} \multicolumn{1}{|l|}{\textbf{PA}} & \textbf{M} & \textbf{Ver} & \multicolumn{1}{l}{\textbf{\#P}} & \multicolumn{1}{c}{\textbf{Cov\%}} & \multicolumn{1}{l}{\textbf{sat}} & \multicolumn{1}{l}{\textbf{us}} & \multicolumn{1}{l}{\textbf{un}} & \multicolumn{1}{l}{\textbf{H}} & \multicolumn{1}{l}{\textbf{HS}} & \multicolumn{1}{l}{\textbf{C(S)}} & \multicolumn{1}{l}{\textbf{C(H)}} & \multicolumn{1}{r}{\textbf{SV {\tiny\VarClock}}} & \multicolumn{1}{r}{\textbf{HV {\tiny\VarClock}}} & \multicolumn{1}{r|}{\textbf{Tot {\tiny\VarClock}}} \\
      \hline
        \multirow{8}[2]{*}{\begin{sideways}Age\end{sideways}} & BM1   & SAT   & 75    & 13.14 & 9     & 58    & 8     & 9     & 1     & .90   & .01   & 14.04 & 11.00 & 25.40 \\
        & \cellcolor[rgb]{ .949,  .949,  .949}BM2 & \cellcolor[rgb]{ .949,  .949,  .949}SAT & \cellcolor[rgb]{ .949,  .949,  .949}141 & \cellcolor[rgb]{ .949,  .949,  .949}26.08 & \cellcolor[rgb]{ .949,  .949,  .949}30 & \cellcolor[rgb]{ .949,  .949,  .949}103 & \cellcolor[rgb]{ .949,  .949,  .949}8 & \cellcolor[rgb]{ .949,  .949,  .949}9 & \cellcolor[rgb]{ .949,  .949,  .949}1 & \cellcolor[rgb]{ .949,  .949,  .949}.85 & \cellcolor[rgb]{ .949,  .949,  .949}.01 & \cellcolor[rgb]{ .949,  .949,  .949}7.76 & \cellcolor[rgb]{ .949,  .949,  .949}5.83 & \cellcolor[rgb]{ .949,  .949,  .949}13.84 \\
        & BM3   & SAT   & 139   & 26.47 & 27    & 108   & 4     & 6     & 2     & .96   & .00   & 9.78  & 3.42  & 13.49 \\
        & \cellcolor[rgb]{ .949,  .949,  .949}BM4 & \cellcolor[rgb]{ .949,  .949,  .949}SAT & \cellcolor[rgb]{ .949,  .949,  .949}37 & \cellcolor[rgb]{ .949,  .949,  .949}6.08 & \cellcolor[rgb]{ .949,  .949,  .949}1 & \cellcolor[rgb]{ .949,  .949,  .949}30 & \cellcolor[rgb]{ .949,  .949,  .949}6 & \cellcolor[rgb]{ .949,  .949,  .949}6 & \cellcolor[rgb]{ .949,  .949,  .949}0 & \cellcolor[rgb]{ .949,  .949,  .949}.91 & \cellcolor[rgb]{ .949,  .949,  .949}.03 & \cellcolor[rgb]{ .949,  .949,  .949}17.10 & \cellcolor[rgb]{ .949,  .949,  .949}16.29 & \cellcolor[rgb]{ .949,  .949,  .949}49.98 \\
        & BM5   & SAT   & 510   & 99.61 & 114   & 394   & 2     & 7     & 5     & .83   & .00   & 2.29  & 0.54  & 2.96 \\
        & \cellcolor[rgb]{ .949,  .949,  .949}BM6 & \cellcolor[rgb]{ .949,  .949,  .949}SAT & \cellcolor[rgb]{ .949,  .949,  .949}510 & \cellcolor[rgb]{ .949,  .949,  .949}100.00 & \cellcolor[rgb]{ .949,  .949,  .949}156 & \cellcolor[rgb]{ .949,  .949,  .949}354 & \cellcolor[rgb]{ .949,  .949,  .949}0 & \cellcolor[rgb]{ .949,  .949,  .949}3 & \cellcolor[rgb]{ .949,  .949,  .949}3 & \cellcolor[rgb]{ .949,  .949,  .949}.76 & \cellcolor[rgb]{ .949,  .949,  .949}.00 & \cellcolor[rgb]{ .949,  .949,  .949}1.17 & \cellcolor[rgb]{ .949,  .949,  .949}0.08 & \cellcolor[rgb]{ .949,  .949,  .949}1.36 \\
        & BM7   & SAT   & 124   & 23.33 & 62    & 57    & 5     & 9     & 4     & .92   & .01   & 8.31  & 5.10  & 14.89 \\
        & \cellcolor[rgb]{ .949,  .949,  .949}BM8 & \cellcolor[rgb]{ .949,  .949,  .949}UNK & \cellcolor[rgb]{ .949,  .949,  .949}23 & \cellcolor[rgb]{ .949,  .949,  .949}3.14 & \cellcolor[rgb]{ .949,  .949,  .949}0 & \cellcolor[rgb]{ .949,  .949,  .949}16 & \cellcolor[rgb]{ .949,  .949,  .949}7 & \cellcolor[rgb]{ .949,  .949,  .949}10 & \cellcolor[rgb]{ .949,  .949,  .949}3 & \cellcolor[rgb]{ .949,  .949,  .949}.79 & \cellcolor[rgb]{ .949,  .949,  .949}.04 & \cellcolor[rgb]{ .949,  .949,  .949}46.23 & \cellcolor[rgb]{ .949,  .949,  .949}31.64 & \cellcolor[rgb]{ .949,  .949,  .949}79.08 \\
  \hline
  \multirow{5}[2]{*}{\begin{sideways}Sex\end{sideways}} & GC1   & SAT   & 31    & 13.43 & 27    & 0     & 4     & 6     & 2     & .76   & .01   & 41.78 & 15.98 & 58.18 \\
        & \cellcolor[rgb]{ .949,  .949,  .949}GC2 & \cellcolor[rgb]{ .949,  .949,  .949}SAT & \cellcolor[rgb]{ .949,  .949,  .949}11 & \cellcolor[rgb]{ .949,  .949,  .949}1.99 & \cellcolor[rgb]{ .949,  .949,  .949}4 & \cellcolor[rgb]{ .949,  .949,  .949}0 & \cellcolor[rgb]{ .949,  .949,  .949}7 & \cellcolor[rgb]{ .949,  .949,  .949}9 & \cellcolor[rgb]{ .949,  .949,  .949}2 & \cellcolor[rgb]{ .949,  .949,  .949}.78 & \cellcolor[rgb]{ .949,  .949,  .949}.04 & \cellcolor[rgb]{ .949,  .949,  .949}97.18 & \cellcolor[rgb]{ .949,  .949,  .949}75.67 & \cellcolor[rgb]{ .949,  .949,  .949}174.18 \\
        & GC3   & SAT   & 201   & 100.00 & 195   & 6     & 0     & 1     & 1     & .69   & .00   & 1.29  & 0.24  & 1.63 \\
        & \cellcolor[rgb]{ .949,  .949,  .949}GC4 & \cellcolor[rgb]{ .949,  .949,  .949}SAT & \cellcolor[rgb]{ .949,  .949,  .949}201 & \cellcolor[rgb]{ .949,  .949,  .949}100.00 & \cellcolor[rgb]{ .949,  .949,  .949}2 & \cellcolor[rgb]{ .949,  .949,  .949}199 & \cellcolor[rgb]{ .949,  .949,  .949}0 & \cellcolor[rgb]{ .949,  .949,  .949}1 & \cellcolor[rgb]{ .949,  .949,  .949}1 & \cellcolor[rgb]{ .949,  .949,  .949}.63 & \cellcolor[rgb]{ .949,  .949,  .949}.00 & \cellcolor[rgb]{ .949,  .949,  .949}0.73 & \cellcolor[rgb]{ .949,  .949,  .949}0.14 & \cellcolor[rgb]{ .949,  .949,  .949}0.97 \\
        & GC5   & UNK   & 12    & 1.49  & 0     & 3     & 9     & 9     & 0     & .59   & .03   & 75.27 & 75.43 & 151.53 \\
  \hline
  \multirow{12}[2]{*}{\begin{sideways}Race\end{sideways}} & AC1   & SAT   & 29    & 0.14  & 8     & 15    & 6     & 8     & 2     & .64   & .04   & 40.73 & 23.50 & 64.43 \\
        & \cellcolor[rgb]{ .949,  .949,  .949}AC2 & \cellcolor[rgb]{ .949,  .949,  .949}SAT & \cellcolor[rgb]{ .949,  .949,  .949}15 & \cellcolor[rgb]{ .949,  .949,  .949}0.04 & \cellcolor[rgb]{ .949,  .949,  .949}4 & \cellcolor[rgb]{ .949,  .949,  .949}3 & \cellcolor[rgb]{ .949,  .949,  .949}8 & \cellcolor[rgb]{ .949,  .949,  .949}10 & \cellcolor[rgb]{ .949,  .949,  .949}2 & \cellcolor[rgb]{ .949,  .949,  .949}.82 & \cellcolor[rgb]{ .949,  .949,  .949}.04 & \cellcolor[rgb]{ .949,  .949,  .949}72.62 & \cellcolor[rgb]{ .949,  .949,  .949}54.86 & \cellcolor[rgb]{ .949,  .949,  .949}128.45 \\
        & AC3   & SAT   & 23    & 0.11  & 16    & 1     & 6     & 10    & 4     & .75   & .02   & 52.75 & 32.33 & 85.40 \\
        & \cellcolor[rgb]{ .949,  .949,  .949}AC4 & \cellcolor[rgb]{ .949,  .949,  .949}UNK & \cellcolor[rgb]{ .949,  .949,  .949}8 & \cellcolor[rgb]{ .949,  .949,  .949}0.00 & \cellcolor[rgb]{ .949,  .949,  .949}0 & \cellcolor[rgb]{ .949,  .949,  .949}0 & \cellcolor[rgb]{ .949,  .949,  .949}8 & \cellcolor[rgb]{ .949,  .949,  .949}8 & \cellcolor[rgb]{ .949,  .949,  .949}0 & \cellcolor[rgb]{ .949,  .949,  .949}.67 & \cellcolor[rgb]{ .949,  .949,  .949}.20 & \cellcolor[rgb]{ .949,  .949,  .949}100.57 & \cellcolor[rgb]{ .949,  .949,  .949}100.08 & \cellcolor[rgb]{ .949,  .949,  .949}241.70 \\
        & AC5   & SAT   & 10    & 0.02  & 3     & 0     & 7     & 9     & 2     & .71   & .13   & 98.14 & 74.06 & 180.41 \\
        & \cellcolor[rgb]{ .949,  .949,  .949}AC6 & \cellcolor[rgb]{ .949,  .949,  .949}SAT & \cellcolor[rgb]{ .949,  .949,  .949}19 & \cellcolor[rgb]{ .949,  .949,  .949}0.07 & \cellcolor[rgb]{ .949,  .949,  .949}6 & \cellcolor[rgb]{ .949,  .949,  .949}5 & \cellcolor[rgb]{ .949,  .949,  .949}8 & \cellcolor[rgb]{ .949,  .949,  .949}9 & \cellcolor[rgb]{ .949,  .949,  .949}1 & \cellcolor[rgb]{ .949,  .949,  .949}.49 & \cellcolor[rgb]{ .949,  .949,  .949}.05 & \cellcolor[rgb]{ .949,  .949,  .949}55.70 & \cellcolor[rgb]{ .949,  .949,  .949}44.43 & \cellcolor[rgb]{ .949,  .949,  .949}100.37 \\
        & AC7   & UNK   & 14    & 0.04  & 0     & 7     & 7     & 10    & 3     & .59   & .11   & 78.22 & 51.09 & 132.28 \\
        & \cellcolor[rgb]{ .949,  .949,  .949}AC8 & \cellcolor[rgb]{ .949,  .949,  .949}SAT & \cellcolor[rgb]{ .949,  .949,  .949}101 & \cellcolor[rgb]{ .949,  .949,  .949}0.63 & \cellcolor[rgb]{ .949,  .949,  .949}82 & \cellcolor[rgb]{ .949,  .949,  .949}19 & \cellcolor[rgb]{ .949,  .949,  .949}0 & \cellcolor[rgb]{ .949,  .949,  .949}1 & \cellcolor[rgb]{ .949,  .949,  .949}1 & \cellcolor[rgb]{ .949,  .949,  .949}.22 & \cellcolor[rgb]{ .949,  .949,  .949}.00 & \cellcolor[rgb]{ .949,  .949,  .949}17.59 & \cellcolor[rgb]{ .949,  .949,  .949}0.13 & \cellcolor[rgb]{ .949,  .949,  .949}17.84 \\
        & AC9   & SAT   & 741   & 4.63  & 399   & 342   & 0     & 4     & 4     & .19   & .00   & 2.29  & 0.02  & 2.44 \\
        & \cellcolor[rgb]{ .949,  .949,  .949}AC10 & \cellcolor[rgb]{ .949,  .949,  .949}SAT & \cellcolor[rgb]{ .949,  .949,  .949}20 & \cellcolor[rgb]{ .949,  .949,  .949}0.09 & \cellcolor[rgb]{ .949,  .949,  .949}6 & \cellcolor[rgb]{ .949,  .949,  .949}8 & \cellcolor[rgb]{ .949,  .949,  .949}6 & \cellcolor[rgb]{ .949,  .949,  .949}10 & \cellcolor[rgb]{ .949,  .949,  .949}4 & \cellcolor[rgb]{ .949,  .949,  .949}.27 & \cellcolor[rgb]{ .949,  .949,  .949}.05 & \cellcolor[rgb]{ .949,  .949,  .949}62.07 & \cellcolor[rgb]{ .949,  .949,  .949}32.82 & \cellcolor[rgb]{ .949,  .949,  .949}95.07 \\
        & AC11  & UNK   & 9     & 0.00  & 0     & 0     & 9     & 9     & 0     & .11   & .02   & 100.13 & 100.12 & 200.66 \\
        & \cellcolor[rgb]{ .949,  .949,  .949}AC12 & \cellcolor[rgb]{ .949,  .949,  .949}UNK & \cellcolor[rgb]{ .949,  .949,  .949}16 & \cellcolor[rgb]{ .949,  .949,  .949}0.04 & \cellcolor[rgb]{ .949,  .949,  .949}0 & \cellcolor[rgb]{ .949,  .949,  .949}7 & \cellcolor[rgb]{ .949,  .949,  .949}9 & \cellcolor[rgb]{ .949,  .949,  .949}9 & \cellcolor[rgb]{ .949,  .949,  .949}0 & \cellcolor[rgb]{ .949,  .949,  .949}.29 & \cellcolor[rgb]{ .949,  .949,  .949}.01 & \cellcolor[rgb]{ .949,  .949,  .949}57.56 & \cellcolor[rgb]{ .949,  .949,  .949}56.35 & \cellcolor[rgb]{ .949,  .949,  .949}114.19 \\
  \hline
      \multicolumn{15}{p{8.9cm}}{\textit{Experiment setup:} soft-timeout 100s, hard-timeout 30m, max-partition size: 100 (BM, GC), 10 (AC). Ver: Verification, \#P: number of partitions, Cov: Coverage, us: UNSAT, un: UNKNOWN, H: \# times heuristic pruning attempted, HS: \# times heuristic pruning succeeded, C: average \textbf{c}ompression ratio-(S): sound pruning, (H): heuristic pruning, Average time (second) {\tiny\VarClock} -- SV: sound verification, HV: heuristic verification.} \\
\end{tabular}%
\label{tab:utility}%
% \vspace{-5mm}
\end{table}%

%% file: stress.tex
\begin{table}[t]
  \centering
  % \footnotesize
  \scriptsize
  \setlength\tabcolsep{1.85pt}
  \caption{Verification with scaled experiment setup}
    \begin{tabular}{|c|l|rrrrrrrrrrr|}
      \hline
      % \rowcolor[rgb]{ .851,  .851,  .851} \multicolumn{1}{|c|}{\textbf{Model}} & \textbf{Ver} & \multicolumn{1}{l|}{\textbf{\#P}} & \multicolumn{1}{l|}{\textbf{\#sat}} & \multicolumn{1}{l|}{\textbf{\#unsat}} & \multicolumn{1}{l|}{\textbf{\#unk}} \\
    \rowcolor[rgb]{ .851,  .851,  .851} \textbf{M} & \textbf{Ver} & \textbf{\#P} & \textbf{\#sat} & \textbf{\#unsat} & \textbf{\#unk} & \textbf{H} & \textbf{HS} & \textbf{C(S)} & \textbf{C(H)} & \textbf{SV {\tiny\VarClock}} & \textbf{HV {\tiny\VarClock}} & \textbf{Total {\tiny\VarClock}} \\
    \hline
    BM8   & SAT   & 48    & 4     & 37    & 7     & 10    & 3     & .85   & .02   & 47.64 & 33.46 & 81.68 \\
    \rowcolor[rgb]{ .949,  .949,  .949} GC5   & UNK   & 12    & 0     & 3     & 9     & 9     & 0     & .60   & .03   & 150.25 & 150.55 & 301.83 \\
    AC4   & SAT   & 24    & 15    & 4     & 5     & 9     & 4     & .92   & .02   & 96.65 & 56.17 & 160.26 \\
    \rowcolor[rgb]{ .949,  .949,  .949} AC7   & UNK   & 21    & 0     & 15    & 6     & 9     & 3     & .77   & .02   & 111.11 & 65.64 & 178.12 \\
    AC11  & UNK   & 9     & 0     & 0     & 9     & 9     & 0     & .21   & .02   & 200.23 & 200.37 & 401.18 \\
    \rowcolor[rgb]{ .949,  .949,  .949} AC12  & SAT   & 35    & 3     & 26    & 6     & 9     & 3     & .41   & .01   & 63.47 & 43.93 & 107.71 \\
    \hline
  \end{tabular}%
  \label{tab:stress}%
  \vspace{3pt}
  Soft-timeout 200s, hard-timeout 60m, max-partition: 10 (for BM, GC), and 6 (for AC)
  % \vspace{-5mm}
\end{table}%

%% file: relaxed.tex
\begin{table*}[htbp]
  \centering
  % \footnotesize
  \scriptsize
  \setlength\tabcolsep{1.3pt}
  \caption{Verification of neural networks for relaxed and targeted fairness queries}
    \begin{tabular}{|c|l|c|rrrrrrrr|c|rrrrr|c|rrrrrrrrrrrr|}
\cline{2-30}    \multicolumn{1}{r|}{} & \textbf{Result} & \multicolumn{1}{c|}{\textbf{$\phi$}} & \textbf{BM1} & \textbf{BM2} & \textbf{BM3} & \textbf{BM4} & \textbf{BM5} & \textbf{BM6} & \textbf{BM7} & \textbf{BM8} & \multicolumn{1}{c|}{\textbf{$\phi$}} & \textbf{GC1} & \textbf{GC2} & \textbf{GC3} & \textbf{GC4} & \textbf{GC5} & \multicolumn{1}{c|}{\textbf{$\phi$}} & \textbf{AC1} & \textbf{AC2} & \textbf{AC3} & \textbf{AC4} & \textbf{AC5} & \textbf{AC6} & \textbf{AC7} & \textbf{AC8} & \textbf{AC9} & \textbf{AC10} & \textbf{AC11} & \textbf{AC12} \\
\hline
    \multirow{10}[3]{*}{\begin{sideways}Relaxed\end{sideways}} & Ver   & \multirow{5}[1]{*}{$\phi_{r11}$} & SAT   & SAT   & SAT   & SAT   & SAT   & SAT   & SAT   & SAT   & \multirow{5}[1]{*}{$\phi_{r12}$} & SAT   & SAT   & SAT   & SAT   & UNK   & \multirow{5}[1]{*}{$\phi_{r13}$} & SAT   & SAT   & SAT   & SAT   & SAT   & SAT   & UNK   & SAT   & SAT   & SAT   & UNK   & UNK \\
          & \#P   &       & 333   & 566   & 1600  & 132   & 4915  & 8671  & 590   & 112   &       & 111   & 32    & 2697  & 9695  & 9     &       & 216   & 75    & 129   & 41    & 37    & 177   & 66    & 639   & 3991  & 216   & 31    & 123 \\
          & \#sat &       & 40    & 65    & 233   & 10    & 731   & 1484  & 168   & 11    &       & 107   & 24    & 2026  & 185   & 0     &       & 9     & 8     & 34    & 5     & 3     & 28    & 0     & 218   & 483   & 19    & 0     & 0 \\
          & \#unsat &       & 279   & 493   & 1358  & 111   & 4180  & 7187  & 411   & 87    &       & 0     & 2     & 671   & 9510  & 0     &       & 193   & 55    & 86    & 22    & 18    & 133   & 53    & 418   & 3508  & 192   & 13    & 107 \\
          & \#unk &       & 14    & 8     & 9     & 11    & 4     & 0     & 11    & 14    &       & 4     & 6     & 0     & 0     & 9     &       & 14    & 12    & 9     & 14    & 16    & 16    & 13    & 3     & 0     & 5     & 18    & 16 \\
\cline{2-30}          & Ver   & \multirow{5}[2]{*}{$\phi_{r21}$} & SAT   & SAT   & SAT   & SAT   & SAT   & SAT   & SAT   & SAT   & \multirow{5}[2]{*}{$\phi_{r22}$} & SAT   & SAT   & SAT   & SAT   & UNK   & \multirow{5}[2]{*}{$\phi_{r23}$} & SAT   & SAT   & SAT   & SAT   & SAT   & SAT   & UNK   & SAT   & SAT   & SAT   & UNK   & SAT \\
          & \#P   &       & 968   & 1514  & 2768  & 157   & 7601  & 13779 & 1340  & 188   &       & 116   & 30    & 3175  & 8361  & 20    &       & 210   & 74    & 124   & 30    & 27    & 117   & 44    & 402   & 4811  & 164   & 18    & 53 \\
          & \#sat &       & 65    & 110   & 208   & 10    & 583   & 1200  & 187   & 6     &       & 106   & 16    & 2343  & 79    & 0     &       & 47    & 20    & 84    & 13    & 11    & 47    & 0     & 182   & 1294  & 34    & 0     & 1 \\
          & \#unsat &       & 890   & 1396  & 2550  & 143   & 7014  & 12579 & 1151  & 167   &       & 1     & 2     & 832   & 8282  & 2     &       & 157   & 45    & 35    & 5     & 3     & 56    & 31    & 220   & 3517  & 128   & 0     & 36 \\
          & \#unk &       & 13    & 8     & 10    & 4     & 4     & 0     & 2     & 15    &       & 9     & 12    & 0     & 0     & 18    &       & 6     & 9     & 5     & 12    & 13    & 14    & 13    & 0     & 0     & 2     & 18    & 16 \\
    \hline
    \multirow{10}[4]{*}{\begin{sideways}Targeted\end{sideways}} & Ver   & \multirow{5}[2]{*}{$\phi_{t11}$} & SAT   & SAT   & SAT   & SAT   & SAT   & SAT   & SAT   & SAT   & \multirow{5}[2]{*}{$\phi_{t12}$} & SAT   & SAT   & SAT   & SAT   & UNK   & \multirow{5}[2]{*}{$\phi_{t13}$} & SAT   & SAT   & SAT   & SAT   & SAT   & SAT   & UNK   & SAT   & SAT   & SAT   & UNK   & UNK \\
          & \#P   &       & 261   & 546   & 1541  & 92    & 5937  & 12233 & 544   & 90    &       & 162   & 32    & 3810  & 11067 & 18    &       & 223   & 67    & 115   & 25    & 23    & 111   & 38    & 448   & 6800  & 108   & 18    & 68 \\
          & \#sat &       & 34    & 61    & 208   & 9     & 881   & 1939  & 157   & 8     &       & 153   & 17    & 2786  & 96    & 0     &       & 55    & 16    & 75    & 7     & 8     & 52    & 0     & 179   & 1624  & 23    & 0     & 0 \\
          & \#unsat &       & 216   & 479   & 1322  & 72    & 5050  & 10294 & 374   & 67    &       & 0     & 2     & 1023  & 10971 & 0     &       & 160   & 37    & 32    & 3     & 1     & 47    & 26    & 266   & 5176  & 81    & 0     & 50 \\
          & \#unk &       & 11    & 6     & 11    & 11    & 6     & 0     & 13    & 15    &       & 9     & 13    & 1     & 0     & 18    &       & 8     & 14    & 8     & 15    & 14    & 12    & 12    & 3     & 0     & 4     & 18    & 18 \\
\cline{2-30}          & Ver   & \multirow{5}[2]{*}{$\phi_{t21}$} & SAT   & SAT   & SAT   & SAT   & SAT   & SAT   & SAT   & SAT   & \multirow{5}[2]{*}{$\phi_{t22}$} & SAT   & SAT   & SAT   & SAT   & UNK   & \multirow{5}[2]{*}{$\phi_{t23}$} & SAT   & SAT   & SAT   & SAT   & SAT   & SAT   & UNK   & SAT   & SAT   & SAT   & UNK   & SAT \\
          & \#P   &       & 432   & 683   & 1987  & 108   & 6474  & 13855 & 986   & 89    &       & 187   & 46    & 5957  & 13426 & 18    &       & 281   & 85    & 119   & 35    & 33    & 119   & 43    & 677   & 6781  & 164   & 18    & 59 \\
          & \#sat &       & 57    & 76    & 270   & 10    & 904   & 2224  & 285   & 11    &       & 175   & 28    & 3534  & 63    & 0     &       & 61    & 18    & 79    & 14    & 16    & 51    & 0     & 309   & 1765  & 36    & 0     & 1 \\
          & \#unsat &       & 361   & 602   & 1705  & 87    & 5569  & 11631 & 693   & 64    &       & 4     & 7     & 2422  & 13363 & 0     &       & 214   & 58    & 33    & 8     & 4     & 57    & 26    & 368   & 5016  & 125   & 0     & 41 \\
          & \#unk &       & 14    & 5     & 12    & 11    & 1     & 0     & 8     & 14    &       & 8     & 11    & 1     & 0     & 18    &       & 6     & 9     & 7     & 13    & 13    & 11    & 17    & 0     & 0     & 3     & 18    & 17 \\
    \hline
    \multicolumn{30}{p{18.3cm}}{\textbf{Relaxed queries:} $\phi_{r11}$: duration < 5, $\phi_{r12}$: job < $\infty$, $\phi_{r21}$: credit-amount < 100, $\phi_{r22}$: foreign-worker < $\infty$, $\phi_{r31}$: marital-status < $\infty$, $\phi_{r32}$: age<5, \textbf{Targeted:} $\phi_{t11}$: personal-loan \& profession:entrepreneur, $\phi_{t12}$: previous-marketing:yes, $\phi_{t21}$: \#credits=2, $\phi_{t22}$: foreign-worker \& credit-purpose:education, $\phi_{t31}$: 30$\le$age$\le$35, $\phi_{t32}$: education:bachelor or doctorate} \\
    % \multicolumn{30}{c}{} \\
    \end{tabular}%
  \label{tab:relaxed}%
  \vspace{-3mm}
\end{table*}%

%% file: threat.tex
\section{Threats to Validity}
\label{sec:threats}

\textbf{Construct validity.} 
% refers to whether the verification results actually provides fairness guarantee of the models. 
We leveraged the fairness definitions from prior work and then encoded using well studied weakest precondition and constraint based methodologies. 
In \secref{subsec:fairness-def}, we argued why individual fairness is useful for verification. However, as mentioned by Corbett-Davies and Goel \cite{corbett2018measure}, the property can fall short in bias quantification. Our evaluation shows that based on the number of certified partitions, future work can be done to quantify individual fairness.
The proposed sound-pruning strategy leverages interval arithmetic which is sound by construction; as shown by [47] i.e., NNs have well-defined transformers (addition, subtraction, scaling) and hence interval analysis provides sound approximation. For individual verification, we used SMT based constraint solving which is always sound. Additionally, input partitioning (\algoref{algo:partition}) creates disjoint partitions of attributes and covers the complete domain.
Furthermore, to be able to verify symbolically, we converted the models into Python functions using the weights and biases extracted from the actual model. We followed the same approach taken by prior work for such conversion \cite{albarghouthi2017fairsquare}.

\textbf{External validity.}
% is concerned about the generalizability of our approach. 
We evaluated Fairify on the popular structured datasets extensively used in prior works \cite{biswas21fair,urban2020perfectly,john2020verifying,albarghouthi2017fairsquare,galhotra2017fairness,aggarwal2019black,udeshi2018automated,bastani2019probabilistic,chakraborty2020fairway,hort2021fairea,chakraborty2021bias,zhang2021ignorance,zhang2020white}. As pointed out by \cite{urban2020perfectly,john2020verifying}, verifying ML models trained on unstructured data including image or natural-language would require different approach. Yet, because of the wide usage in real-world (loan approval, criminal sentencing, etc.), this line of work only considered such structured data. 
In addition, followed by \cite{zhang2020white,urban2020perfectly,mazzucato2021reduced,aggarwal2019black,udeshi2018automated}, we considered the ReLU based NNs. To further demonstrate the applicability, we collected top-rated models from practice, which are state-of-the-art accurate models for respective tasks. 
Verifying fairness of other classes of NNs such as CNN or LSTM could be potential future improvements.
Finally, Fairify is built on top of the popular open source libraries and SMT solver Z3 so that other works can leverage the tool.

%% file: related.tex
\section{Related Works}
\label{sec:related}

\textit{\textbf{Fairness Testing and Verification.}} 
With the increasing need to ensure fairness of AI based systems \cite{biswas20machine, biswas21fair, zhang2021ignorance, hort2021fairea, chakraborty2020fairway}, many recent works focused on the fairness testing and verification of ML models \cite{galhotra2017fairness, udeshi2018automated, aggarwal2019black, zhang2020white, albarghouthi2017fairsquare, bastani2019probabilistic, john2020verifying,zheng2021neuronfair}. 
While many of the prior works focused on testing ML models \cite{galhotra2017fairness, udeshi2018automated, aggarwal2019black}, more recent works proposed individual fairness testing on NN \cite{zheng2021neuronfair,zhang2020white}. While input test generation has been helpful to find fairness violations, verification is more difficult since it proves the property.

Probabilistic verification techniques have been proposed to verify group fairness property \cite{albarghouthi2017fairsquare,bastani2019probabilistic}. 
Recently, \citeauthor{john2020verifying} proposed individual fairness verification approach for two different kinds of ML models \cite{john2020verifying}, which do not apply for NN.
\citeauthor{urban2020perfectly} proposed Libra to provide certification of another property called dependency fairness using abstract interpretation \cite{urban2020perfectly}. 
% They used forward and backward analysis to compute input region that is fair or unfair using different abstract domains e.g., Boxes, Symbolic, Deeppoly. 
\citeauthor{mazzucato2021reduced} extended Libra with one abstract domain \cite{mazzucato2021reduced}. % and implemented automated configuration
Another class of works proposed methods of individually fair learning by enforcing it during model training \cite{ruoss2020learning,yurochkin2019training}. However, our focus in this paper is to verify the already trained models in production.

\textit{\textbf{NN Verification.}} 
Verification of neural network has been studied for various application domains and property of interest. The robustness of NN has gained a lot of attention for safety-critical applications e.g., autonomous vehicles \cite{nguyen2015deep, chaudhuri2012continuity, goodfellow2014explaining}. Research showed that NN can be fooled by applying small perturbations to data instances i.e., adversarial inputs \cite{szegedy2014google, nguyen2015deep, carlini2017towards, goodfellow2014explaining, zhang2020deepsearch}. Algorithms have been proposed to detect adversarial inputs and satisfy local robustness property \cite{athalye2018obfuscated,carlini2017towards, engstrom2018evaluating, huang2015learning,katz2017towards}. 
% Other research tackled NN verification for robustness \cite{huang2017safety, gopinath2018deepsafe,barrett2018satisfiability}.
\secref{sec:fairness-verification} further describes NN verification and how it compares with the fairness property.
\citeauthor{katz2017reluplex} proposed efficient SMT solving algorithms to provide robustness guarantee in NN \cite{katz2017reluplex, katz2019marabou}. In addition, some verification algorithms leverage off-the-shelve SMT solver \cite{ehlers2017formal, huang2017safety}. Research has also been conducted to compute bounds of the neurons and provide probabilistic guarantees for some properties \cite{sun2018concolic, wang2018efficient, singh2019abstract}. With the extensive use of NN, many recent works focused on new types of properties using both static and dynamic analysis techniques \cite{paulsen2020reludiff, carlini2017towards, ma2018mode, tian2018deeptest, huchard2018proceedings}.
%\citeauthor{xiao2021self} showed that the decision of early layers can be used to give warning \cite{xiao2021self}. 
% Finally, many prior works focused on other dynamic techniques and testing methodologies to suggest violation of safety properties \cite{carlini2017towards, ma2018mode, tian2018deeptest, huchard2018proceedings}.

%% file: conclusion.tex
\section{Conclusion and Future Work}
\label{sec:conc}

In this work, we addressed the fairness verification problem of neural networks. 
% Given the complex structure, providing formal guarantee to the NN is difficult.
% Previous works considered NN certification for different properties. 
% We are the first to verify individual fairness property for NN that has been shown to be useful in fairness testing in prior work. 
Our technique, Fairify can verify individual fairness and its relaxations on real-world NN models, which has not been tackled before.
We proposed lightweight techniques to reduce the problem into multiple sub-problems and prune the networks to reduce the verification complexity. Fairify applies static interval analysis and individual verification to provably prune the neurons. In addition, we conducted neuron profiling to observe their heuristic and prune further.
While many prior works focused on individual fairness testing and improving, Fairify provides formal guarantee of fairness. Our work also bridges the gap between the theoretical formal analysis and its usage in real-world, as Fairify provides several practical benefits for the developers, e.g., provide counterexample, targeted fairness.
% Fairify showed much improvement over the baseline in verifying many real-world NN. 
The result of Fairify can be leveraged in fairness testing for guided test case generation.
%The pruned NN has added benefit since it can be deployed in the production.
Also, the counterexamples can be used to repair the NN in interactive verification setting.
Novel analysis can be proposed in the future to prioritize the verification of partitions and dynamically allocate time. Fairify would also be leveraged provable repair and design of fairness-aware software.